\documentclass[final]{ecai}

\usepackage{amsfonts}
\usepackage{amsmath}
\usepackage[utf8]{inputenc}
\usepackage{booktabs}
\usepackage{tabu}
\usepackage{arydshln}
\usepackage[inline]{enumitem}
\usepackage[T1]{fontenc}
\usepackage{graphicx}
\usepackage{latexcolors}
\usepackage{multirow}
\usepackage{nicefrac}
\usepackage{xcolor}
\usepackage{tikz}
\usetikzlibrary{arrows.meta, matrix, positioning, shapes}

\usepackage{algorithm}
\usepackage{algorithmic}
\usepackage{tabu}
\usepackage{multirow}

\newcommand{\I}{\mathcal{I}}
\newcommand{\B}{\mathcal{B}}
\newcommand{\D}{\mathcal{D}}
\newcommand{\X}{\mathcal{X}}

\definecolor{cb-yellow}{HTML}{F5C710}
\definecolor{cb-green}{HTML}{009E73}
\definecolor{cb-blue}{HTML}{56B4E9}
\definecolor{cb-vermilion}{HTML}{D55E00}

\tikzset
{
    every node/.style =
    {
        align = center
    }
}

\tikzset
{
    every path/.style =
    {
        -stealth,
        draw,
    }
}

\tikzset
{
    function/.style = 
    {
        rectangle,
        minimum size = 0.75cm,
        font = \footnotesize,
        align = center,
        fill = cb-green,
        draw = white,
        text = white,
        line width = 0cm
    }
}

\tikzset
{
    dataset/.style = 
    {
        cylinder,
        minimum size = 2cm,
        font = \footnotesize,
        shape border rotate = 90,
        shape aspect = 0.4,
        align = center,
        fill = cb-blue,
        draw = white,
        text = white,
        line width = 0cm
    }
}

\tikzset
{
    algorithm/.style =
    {
        diamond,
        minimum size = 0.75cm,
        font = \footnotesize,
        align = center,
        fill = cb-vermilion,
        draw = white,
        text = white,
        line width = 0cm
    }
}

\tikzset
{
    preference/.style =
    {
        cylinder,
        minimum width = 1.75cm,
        minimum height = 1.5cm,
        font = \footnotesize,
        shape border rotate = 90,
        shape aspect = 0.25,
        align = center,
        fill = cb-blue,
        draw = white,
        text = white,
        line width = 0cm
    }
}

\tikzset
{
    instance/.style = 
    {
        cylinder,
        minimum width = 1.5cm,
        minimum height = 1cm,
        font = \footnotesize,
        shape border rotate = 90,
        shape aspect = 0.25,
        align = center,
        fill = cb-blue,
        draw = white,
        text = white,
        line width = 0cm
    }
}

\tikzset
{
    prediction/.style =
    {
        circle,
        minimum size = 0.75cm,
        font = \footnotesize,
        fill = cb-yellow,
        draw = white,
        text = white,
        line width = 0cm
    }
}

\tikzset
{
    annotation/.style =
    {
        align = center,
        font = \footnotesize
    }
}

\begin{document}

\begin{frontmatter}

\paperid{4636}

\title{A comparative analysis of rank aggregation methods for the partial label ranking problem}

\author[A]
{%
    \fnms{Jiayi}~\snm{Wang}%
    \orcid{0009-0001-5701-7351}%
}

\author[B,C]
{%
    \fnms{Juan C.}~\snm{Alfaro}%
    \orcid{0000-0003-1777-8540}%
    \thanks{Corresponding Author. Email: JuanCarlos.Alfaro@uclm.es}%
}

\author[D]
{%
    \fnms{Viktor}~\snm{Bengs}%
    \orcid{0000-0001-6988-6186}%
}

\address[A]
{%
    Chair of Artificial Intelligence and Machine Learning, %
    Ludwig-Maximilians-Universität München, %
    Munich, Germany%
}

\address[B]
{%
    Departamento de Sistemas Informáticos, %
    Universidad de Castilla-La Mancha, %
    Albacete, Spain%
}

\address[C]
{%
    Laboratorio de Sistemas Inteligentes y Minería de Datos, %
    Universidad de Castilla-La Mancha, %
    Albacete, Spain%
}

\address[D]
{%
    German Research Center for Artificial Intelligence, 
    Kaiserslautern, Germany%
}

\begin{abstract}
    The \textit{label ranking} problem is a supervised learning scenario in which the learner predicts a \textit{total order} of the class labels for a given input instance. 
    Recently, research has increasingly focused on the \textit{partial label ranking} problem, a generalization of the label ranking problem that allows \textit{ties} in the predicted orders.
    So far, most existing learning approaches for the partial label ranking problem rely on approximation algorithms for rank aggregation in the final prediction step.
    This paper explores several alternative aggregation methods for this critical step, including scoring-based and non-parametric probabilistic-based rank aggregation approaches.
    To enhance their suitability for the more general partial label ranking problem, the investigated methods are extended to increase the likelihood of producing ties.
    Experimental evaluations on standard benchmarks demonstrate that scoring-based variants consistently outperform the current state-of-the-art method in handling incomplete information. In contrast, non-parametric probabilistic-based variants fail to achieve competitive performance.
\end{abstract}

\end{frontmatter}

\section{Introduction}
\label{section:introduction}

The rise of \textit{large language models} has significantly heightened interest in \textit{preference learning}.
This field explores preference information in various forms, including \textit{absolute} nature such as \textit{utility scores}, \textit{binary}, or \textit{ordinal data}, as well as \textit{relative} nature like \textit{total orders}, \textit{top-k lists}, or \textit{partial orders}.
This attentiveness arises from the significant contributions of preference learning techniques to optimizing the fine-tuning process of large language models, as discussed in recent surveys on the subject \cite{kaufmann_survey_2024}. 
This renewed interest encompasses a broad range of topics, from fundamental areas such as \textit{statistical preference models} \cite{firth_davidson-luce_2019, henderson_modelling_2024} to advanced domains like \textit{preference-based bandits} \cite{bengs_preference-based_2021} or \textit{preference-based reinforcement learning} \cite{wirth_survey_2017}.

Another important area within this range is \textit{label ranking}, which focuses on the instance-dependent prediction of total orders over a set of predefined class labels \cite{hullermeier_label_2008}.
For example, consider a movie recommendation task where each instance is a user, described by features such as age or watch history.
The goal is to predict a total order over predefined movie genres (e.g., action, comedy, drama), reflecting the user's preferences from most to least preferred.

Although there is extensive literature on this learning problem, its applicability to real-world scenarios remains limited. 
For example, when predicting movie popularity, label ranking methods inherently produce a total ordering of movies.
However, it is questionable whether every user can consistently rank each pair of movies as being either \textit{more} or \textit{less preferred}.
This limitation is not restricted to typical recommendation system domains like movies or music; it also arises in broader scenarios involving human preferences.
A notable example is the preferences among different responses generated by a large language model to a user's prompt.

A straightforward way to address the limitations of predicting a total order is introducing a third type of relation between class labels: \textit{equally preferred}.
Rather than forcing a total order, this relation allows the expression of uncertainty or genuine indifference between labels.
This concept is central to the \textit{partial label ranking} problem, which generalizes the traditional label ranking framework by using a \textit{bucket order} as its prediction target instead of total orders \cite{alfaro_learning_2021}.
In a bucket order, class labels can be partially ordered, with one or more class labels grouped into a \textit{bucket} to represent \textit{equal preference}.

In the label ranking problem, learning typically focuses on modeling pairwise relationships between class labels by estimating the probability that one label is preferred over another.
Interestingly, many partial label ranking methods adopt this same strategy, using pairwise preference probabilities as their learning objective.
However, unlike in the label ranking problem, where these pairwise scores are aggregated into a total order, partial label ranking aggregates them into a more flexible structure that allows ties, such as a bucket order.
This difference in the aggregation target, total versus bucket order, marks a key distinction between the label ranking problem and its generalization, the partial label ranking problem.
In both settings, the transformation from pairwise preferences to a final prediction relies on well-established \textit{rank aggregation} techniques, which play a central role in preference learning.

So far, the variety of aggregation methods employed in partial label ranking remains limited, primarily relying on the \textit{optimal bucket order} problem \cite{gionis_algorithms_2006, ukkonen_randomized_2009}.
This problem generalizes the well-known \textit{Kemeny consensus} \cite{kemeny_mathematics_1959} used in label ranking, which seeks a total order minimizing pairwise disagreements.
In contrast, the optimal bucket order problem allows ties by producing a bucket order, making it better suited to partial label ranking scenarios where indifference between class labels is common.
This reliance is surprising because both are NP-hard problems, often requiring complex approximation techniques.
Naturally, this complexity influences the learning and inference processes of all methods built around these rank aggregation approaches.

This paper aims to reintroduce various rank aggregation methods to the research community, as outlined in Section \ref{section:aggregations}, and systematically assess their suitability for partial label ranking problems.
The range of methods investigated spans from simple, \textit{scoring-based} approaches to more advanced, \textit{non-parametric probabilistic-based} methods.
A key contribution of this work is adapting certain ranking aggregation methods to address partial label ranking scenarios effectively.
Our experimental evaluation, presented in Section \ref{section:experiments} and conducted using standard benchmarks, reveals that scoring-based variants generally outperform the current state-of-the-art when dealing with incomplete information, while non-parametric probabilistic-based methods lag behind.
As a side product, we propose a technique to obtain suitable hyperparameter values for the extensions of the suggested simple aggregation methods based on the dataset's metadata.
The paper begins with a brief review of related literature in the next section, followed by a concise introduction to the partial label ranking problem in Section \ref{section:preliminaries}.

\section{Related work}
\label{section:related}

Rank aggregation is one of the most pervasive yet underappreciated topics in preference learning. 
Its ubiquity stems from its foundational role, as many concepts and techniques of rank aggregation are integral to framing either the learning problem or the learning methodology across all areas of preference learning.
Its significance is comparable to measures of central tendency in classical probability theory, where measures like the mean or expected value serve as core principles. 
Moreover, rank aggregation is critical in numerous machine learning problems, often without being the primary focus, despite its fundamental impact on the overall learning framework.

Similar to how the mean, or expected value, often serves as a gold standard in classical probability theory, the Kemeny consensus plays a similar role in learning problems involving preferences without ties \cite{brinker_case-based_2006, jiao_controlling_2016, korba_structured_2018}. 
When ties are permitted, the optimal bucket order problem becomes the corresponding standard, as it remains the primary rank aggregation problem used for partial label ranking algorithms \cite{alfaro_learning_2021, alfaro_mixture-based_2021, alfaro_ensemble_2023, alfaro_multi-dimensional_2023, alfaro_pairwise_2023}, except for the approach in \cite{thies_more-plr_2024}.
This preference for the Kemeny consensus and the optimal bucket order problem largely stems from their natural definitions: both aim to identify the object, whether a total order or bucket order, that minimizes the distance to the underlying data among all possible objects.

However, this straightforward formulation has a significant drawback: computing a Kemeny consensus is NP-hard \cite{bartholdi_computational_1989}, as is solving the optimal bucket order problem \cite{lorena_biased_2021}.
To address these computational challenges, several approximation algorithms have been developed \cite{aledo_utopia_2017, aledo_approaching_2019, aledo_highly_2021, gionis_algorithms_2006, ukkonen_randomized_2009}.
An alternative strand of research in preference learning investigates the use of other aggregation methods for the underlying learning problem \cite{bengs_preference-based_2021, hullermeier_preference_2024}.

These methods generally fall into two categories:
\begin{itemize}

    \item \textit{Scoring-based} approaches, such as the \textit{Borda} method \cite{black_partial_1976} or \textit{Copeland} scores \cite{copeland_reasonable_1951}.

    \item \textit{Probabilistic-based} approaches, which include \textit{parametric} methods, like \textit{maximum likelihood estimator} or \textit{Bayes estimator} under assumed probabilistic models \cite{cheng_label_2010}, and \textit{non-parametric} methods, such as \textit{Markov chain-based} approaches \cite{dwork_rank_2001} or modified \textit{von-Neumann winner} concepts \cite{dudik_contextual_2015}.

\end{itemize}
Recently, two label ranking approaches have been proposed that fall outside the two primary categories mentioned above: one based on imprecise probabilities as the core mathematical tool for rank aggregation \cite{adam_inferring_2024} and another employing heuristic search methods \cite{zhou_heuristic_2024}.

Finally, it is worth mentioning the close connection between rankings and tournament solutions. Several works in the field of preference learning explicitly leverage this relationship to translate between the two problems \cite{ramamohan_dueling_2016}.

\section{Preliminaries} 
\label{section:preliminaries}

This section presents the foundational framework for the partial label ranking problem, introducing key concepts such as rankings and the optimal bucket order problem.

\subsection{Ranking}

A \textit{ranking} represents a \textit{preference relation} among a set of \textit{items} \(\I = \{1, \dots, n\}\). 
Following standard conventions in preference learning, the pairwise preference relations between two items \(u, v \in \I\) are denoted as \(u \succ v\) if \(u\) is preferred over \(v\), \(u \prec v\) if \(v\) is preferred over \(u\), and \(u \sim v\) if there is no preference between \(u\) and \(v\).

Rankings can be distinguished along two important dimensions: \textit{completeness} and the \textit{presence of ties}.
Regarding completeness, a ranking can be categorized as either \textit{complete} or \textit{incomplete}, depending on whether a preference relation exists for each pair of items \(u, v \in \I\). 
Similarly, concerning the presence of ties, rankings can be either \textit{with} or \textit{without ties}.

A complete ranking with ties is represented by a \textit{bucket order}.
A bucket order \(\B\) over \(\I\) is an ordered partition of \(k\) disjoint subsets \(\B_{1}, \dots, \B_{k}\), called \textit{buckets}, where \(1 \le k \le n\) and \(\cup _{i = 1}^{k}\B_i = \I\). 
The relations \(\succ_{\B}\), \(\prec_{\B}\), and \(\sim_{\B}\) express preferences between buckets; for example, for two buckets \(\B_i, \B_j \in \B\), \(\B_i \succ_{\B} 
\B_j\) means that \(\B_i\) precedes \(\B_j\) in the bucket order \(\B\). 

The structure of a bucket order naturally defines \textit{pairwise preference relations}.
For \(u \in \B_i\), the bucket \(\B_i\) is referred to as the bucket of \(u\).
Within each bucket \(\B_i\), all items are assumed to be \textit{tied} or \textit{incomparable}, expressed as \(u \sim v\) for \(u, v \in \B_i\). 
Consequently, given \(u \in \B_i\) and \(v \in \B_j\), with \(\B_i \succ_{\B} \B_j\), it follows that \(u \succ v\).
%
%

\subsection{Optimal bucket order problem}
\label{section:obop}

Let us introduce some concepts before delving into the \textit{optimal bucket order} problem.
A \textit{bucket matrix} \(B\) is an \(n \times n\) square matrix associated with a bucket order \(\B\).
Each entry \(B(u, v)\), for \(u, v \in \I\), is defined based on the preference relation in \(\B\) as follows: \(B(u, v) = 1\) if \(u \succ_{\B} v\), \(B(u, v) = 0.5\) if \(u \sim_{\B} v\), and \(B(u, v) = 0\) if \(u \prec_{\B} v\).

A \textit{pair order matrix} \(C\) is an \(n \times n\) square matrix used to represent pairwise preference relations numerically.
For items \(u, v \in \I\), the entry \(C(u, v)\) quantifies the probability of \(u \succ v\), where \(C(u, v) \in [0, 1]\). 
This pairwise relation adheres to \textit{symmetry} and \textit{reflexivity} properties: for all \(u, v \in \I\) with \(u \neq v\), \(C(u, v) + C(v, u) = 1\), and \(C(u, u) = 0.5\).
In practice, the pair order matrix is typically estimated from a collection of (possibly incomplete) partial rankings over the item set; see \cite{alfaro_learning_2021} for further details.

The pair order matrix encodes pairwise preferences, each being an intuitive and obvious prediction object.
However, the entire pair order matrix does not directly correspond to a bucket order over all items and is not readily usable for predictions in the partial label ranking problem.
To address this, a \textit{rank aggregation} problem must be solved to transform the pair order matrix into a \textit{consensus bucket order}.
The most prominent methods consider the optimal bucket order problem.

In the optimal bucket order problem, the objective is to compute the bucket matrix \(B\) that best captures the data represented in the pair order matrix \(C\); in other words, to find a bucket matrix \(B\) that is closest to the pair order matrix \(C\). 
Formally, given a pair order matrix \(C\), the goal is to find a bucket matrix \(B\) which minimizes the \(L^1\) distance \(D(B, C)\) between them. This distance is defined as:
\begin{center}
   $ D(B, C) = \sum\nolimits_{u, v\in \I} \left|B(u, v) - C(u, v) \right|. $
\end{center}

Since the optimal bucket order problem is NP-hard, several approximation methods have been proposed to tackle it \cite{aledo_utopia_2017, aledo_approaching_2019, aledo_highly_2021, gionis_algorithms_2006, ukkonen_randomized_2009}.
A foundational approach is the \textit{bucket pivot} algorithm \cite{gionis_algorithms_2006, ukkonen_randomized_2009}, which serves as the basis for more refined strategies. 
One such refinement is the \textit{least indecision assumption} algorithm \cite{aledo_utopia_2017}, which selects as pivot the item in \(\I\) with the lowest degree of imprecision and incorporates improvements such as \textit{two-stage} and \textit{multi-pivot} strategies.
This technique has become the current state-of-the-art method for the optimal bucket order problem, striking a balance between accuracy and computational efficiency, with a runtime of \(\mathcal{O}(n \log n)\).
%
%

\subsection{Partial label ranking problem}

The \textit{partial label ranking} problem can be framed as a \textit{non-standard supervised classification} problem, in which a \textit{preference model} is trained on a labeled dataset and subsequently used to map instances to bucket orders \cite{vembu_label_2010}.
Figure \ref{figure:plr} illustrates the complete learning and inference process in the partial label ranking problem.

\begin{figure*}[t]
    \centering
    \resizebox{15cm}{!}
{
    \begin{tikzpicture}[thick]

        \node (dataset) [dataset] {Dataset};

        \node (algorithm) [algorithm, right = 2cm of dataset] {\(\mathcal{A}\)};
        \path (dataset) edge node [annotation, above] {Used by} (algorithm);
        \node [annotation, below = 0.1cm of algorithm] {Partial label\\ranking algorithm};

        \node (model) [function, right = 2cm of algorithm] {\(\mathcal{M}\)};
        \path (algorithm) edge node [annotation, above] {Train} (model);
        \node [annotation, below = 0.1cm of model] {Partial label\\ranker};

        \node (instance) [instance, above = 2cm of model] {Instance};
        \path (instance) edge node [annotation, above, sloped] {Use} (model);

        \node (matrix) [preference, right = 2cm of model] {Pair order\\matrix};
        \path (model) edge node [annotation, above] {Fill} (matrix);

        \node (aggregation) [algorithm, right = 2cm of matrix] {\(\mathcal{A}\)};
        \path (matrix) edge node [annotation, above] {Input} (aggregation);
        \node [annotation, below = 0.1cm of aggregation] {Rank aggregation\\algorithm};

        \node (prediction) [prediction, right = 2cm of aggregation] {\(\B\)};
        \path (aggregation) edge node [annotation, above] {Aggregate} (prediction);
        \node [annotation, below = 0.1cm of prediction] {Consensus bucket\\order};

    \end{tikzpicture}
}
    \caption{Example of the complete learning and inference process in the partial label ranking problem}
    \label{figure:plr}
\end{figure*}

In the initial phase, a training dataset is provided, typically collected by asking individuals to rank items based on specific, predefined criteria.
Each instance in this dataset represents a distinct entity characterized by a set of attributes.
Alongside these attributes, each instance conveys a preference for the items, expressed as a (possibly incomplete) ranking with ties.
A partial label ranking algorithm is then used to train a preference model, referred to as a \textit{partial label ranker}.
Given an input instance, this preference model predicts a pair order matrix, which is then processed by a rank aggregation algorithm.
This produces the consensus bucket order, which serves as the prediction and constitutes the final step in the inference phase.

In partial label ranking literature, the consensus bucket order is typically obtained by solving the optimal bucket order problem.
For this, the current state-of-the-art method, as described in Section \ref{section:obop}, is employed.
Consequently, this algorithm is used as a baseline for comparing the performance of alternative aggregation techniques.

Formally, the partial label ranking problem involves a finite set of \(n\) items to be ranked \(\I = \{1, \dots, n\}\) and a set of \(N\) training pairs \(\D = \{(x^i, \pi^i)\}_{i = 1}^{N} \subseteq \mathcal{X} \times \mathcal{S}_\I^{\succeq, \mathrm{inc}}\).
Here, \(\X\) denotes the instance space (typically a subset of \(\mathbb{R}^m\)) and \(\mathcal{S}_\I^{\succeq, \mathrm{inc}}\) denotes the set of (possibly incomplete) partial rankings defined over \(\mathcal{I}\).
Each pair \((x^i, \pi^i) \in \D\), consists of a configuration of values over \(m\) predictive features, i.e., \(x^i = (x_1^i, \dots, x_m^i)\), and a (possibly incomplete) ranking with ties of the values in \(\I\), i.e., \(\pi^{i}\). 
Given these, a partial label ranking algorithm learns a preference model, which is then utilized to assign a bucket order \(\B\) to any input instance \(x \in \X\).
%
%

\section{Alternative rank aggregation algorithms}
\label{section:aggregations}

This section explores alternative rank aggregation methods designed to address the partial label ranking problem, providing different approaches to enhance the prediction of consensus bucket orders. Pseudocode for each algorithm is provided in the supplementary material \cite{wang_comparative_2025}.

\subsection{Scoring-based}
\label{section:scoring}

In \textit{scoring-based} aggregation algorithms, the primary focus is on the values within the given pair order matrix \(C\). 
Specifically, a scoring function \(S: \I \to \mathbb{R}\) assigns a score \(S(u)\) to each item \(u\). 
For most scoring functions, the score \(S(u)\) depends on the pairwise preference relations between an item \(u\) and every other item in \(\I\).

Once the scores for each item have been calculated, the items are ranked according to their respective scores, following a clear value-based ranking criterion.
Items are ranked in decreasing order of their scores.
For instance, if the score of one item \(u\) is larger than that of another item \(v\), i.e., \(S(u) > S(v)\), then \(u\) is ranked ahead of \(v\), i.e., \(u \succ v\).
On the other hand, if the values are the same, i.e., \(S(u) = S(v)\), then \(u\) and \(v\) are considered tied, i.e., \(u \sim v\), and are placed in the same bucket within the bucket order.
The value-based ranking criteria are precise, ensuring no ambiguous rankings and guaranteeing that a bucket order is always present in the results.

Scoring-based aggregation algorithms are often preferred for rank aggregation due to their ability to balance accuracy and computational efficiency. 
Next, we describe two well-known scoring-based aggregation algorithms: \textit{Copeland} and \textit{Borda}.
%
%

\subsubsection*{Copeland}

The \textit{Copeland} aggregation algorithm defines its scoring function \(S\) based on the total number of wins for each item in pairwise comparisons \cite{copeland_reasonable_1951}. 
Specifically, an item \(u\) receives one point for winning a pairwise comparison against \(v\) if \(C(u, v) > 0.5\).
Nevertheless, a key drawback of this approach is that it does not account for the margin of victory.
To be more precise, even when \(C(u, v) - C(v, u) \approx \epsilon\) for a small \(\epsilon > 0\), \(u\) still receives a full point for winning~over~\(v\).

To address this issue, we introduce a hyperparameter \(\beta \geq 0\) to control the threshold for determining a clear winner.
With \(\beta\), item \(u\) earns one point for winning a pairwise comparison against \(v\) only if \(C(u, v) > 0.5 + \beta\).
In cases where the scores are very close, resulting in a draw within the range \(0.5 - \beta \leq C(u, v) \leq 0.5 + \beta\), \(u\) earns half a point.
Conversely, \(u\) receives no points if \(C(u, v) < 0.5 - \beta\).

Formally, the scoring function \(S\) for an item \(u\) in our adaptation of the Copeland algorithm is computed as:
\begin{equation*}
S(u) = \sum_{v \in \I:v\neq u}
    \begin{cases}
        1 & \text{ if } C(u, v) > 0.5 + \beta, \\
        0.5 & \text{ if } 0.5 + \beta \geq C(u,v) \geq 0.5 - \beta, \\
        0 & \text{ if } C(u,v) < 0.5 - \beta.
    \end{cases}
\end{equation*}
Note that introducing the hyperparameter increases the likelihood of producing ties or buckets with the modified Copeland aggregation method.
This, in turn, enhances its suitability for the more general partial label ranking problem.
%
%

\subsubsection*{Borda}

The fundamental concept of the \textit{Borda} aggregation algorithm is, roughly speaking, by ranking items by their average probability of being preferred over another item chosen uniformly at random \cite{black_partial_1976}.
Formally, the scoring function \(S\) in Borda is thus defined as:
\begin{equation*}
    S(u) = \frac{1}{n-1}\sum\nolimits_{v \in \I:v\neq u} C(u, v).
\end{equation*}
For the sake of convenience, we will omit the normalizing factor \(\nicefrac{1}{n-1}\), as this does not change the ordering.

The main problem with this approach lies in the sorting procedure: items are assigned the same rank if and only if they have the exact same score.
This strict criterion makes it unlikely for items to be grouped in the same bucket, even when their scores are close, often resulting in bucket orders that are nearly total orders.
To mitigate this issue, we once again introduce a hyperparameter \(\beta \geq 0\) for this issue in order to relax the grouping criterion. 
The idea is to place items in the same bucket if the difference between their scores falls within the range defined by \(\beta\).
Specifically, for two items \(u\) and \(v\), if \(S(u) - S(v) \in [-\beta, \beta]\), then \(u \sim v\).
This mechanism naturally extends to multiple items through a \textit{cascading effect}, ensuring that items with closely related scores are consistently placed into the same bucket.

It is worth pointing out that the computational complexity for scoring from the pair order matrix is \(\mathcal{O}(n^{2})\) for both scoring-based algorithms.
On the other hand, they exhibit \(O(n\log n)\) complexity in the sorting phase.
Finally, it is worth mentioning that both classical variants of the Copeland or the Borda aggregation methods can be recovered by setting the introduced hyperparameter \(\beta\) to zero.
%

\subsection{Probabilistic-based}
\label{section:probabilistic}

In the following, we revisit several common non-parametric probabilistic approaches to rank aggregation. We exclude parametric methods, such as those based on statistical ranking models like \textit{Plackett-Luce} \cite{plackett_analysis_1975} or \textit{Mallows} \cite{mallows_non-null_1957}, as they do not natively support ties.

\subsubsection*{Markov chain}
\label{section:markov}

Four \textit{Markov chain} algorithm variants have been developed for rank aggregation \cite{dwork_rank_2001}.
The primary objective of these algorithms is to generate an aggregated ranking that minimizes the influence of items that are spuriously ranked highly in only a minority of lists. 

The rationale behind using Markov chain algorithms for rank aggregation is that each item is treated as a state in a \textit{random walk}, where transition probabilities reflect pairwise preference strengths.
The random walk generally moves from less preferred items to more preferred ones.
Running the process until convergence obtains a stationary distribution, the value of which at each state reflects the item's consensus score.
The irreducibility and aperiodicity of the Markov chain ensure convergence.

The algorithms begin by representing a Markov chain with undirected edges.
Each Markov chain is defined by an \(n \times n\) \textit{transition matrix} \(P\), where \(P(u, v)\) represents the transition probabilities from state \(u\) to state \(v\).
The transitioning mechanism varies across the different algorithms, that is, the pattern of state transitions in the constructed Markov chain is not uniform.
In particular, upon reaching state \(u\), the next state \(v\) is selected as follows:
\begin{itemize}
    \item in \textit{Markov chain 1}, uniformly at random from the set of items ranked at least as high as \(u\) in at least one input ranking;
    \item in \textit{Markov chain 2}, by first picking a random input ranking and then choosing \(v\) uniformly at random from the set of items ranked at least as high as \(i\) in that ranking;
    \item in \textit{Markov chain 3}, by choosing a random item \(v\) from a random input ranking;
    \item in \textit{Markov chain 4}, uniformly at random.
\end{itemize}

One limitation of the first three Markov chain algorithms is that their transition matrix computations rely on a set of rankings and cannot be derived directly from the pair order matrix (see \cite{freund_rank_2019} for detailed formulations).
In contrast, Markov chain 4 allows the direct computation of each entry \(P(u, v)\) in the transition matrix from a given pair order matrix \(C\), using the following formulation:
\begin{equation*}
    P(u, v) =
        \begin{cases}
                \nicefrac{1}{n} & \text{ if } u \neq v \land C(u, v) \leq 0.5, \\
                0 & \text{ if } u \neq v \land C(u, v) > 0.5, \\
                1 - \sum_{w:w \neq u} P(u, v) & \text{ if } u = w.
        \end{cases}
\end{equation*}
This direct computation provides an advantage for Markov chain 4 when only a pair order matrix is available \cite{alfaro_multi-dimensional_2023, alfaro_pairwise_2023}.
To evaluate these algorithms, we employed a partial label ranker capable of predicting both the pair order matrix and the subset of (possibly incomplete) partial rankings required to derive it.

Once the Markov chain with the transition matrix is constructed, the algorithms will run on this chain until convergence.
Subsequently, the last step is to find the stationary distribution \(x\) of this Markov chain that satisfies \(x = xP\).

After getting the stationary distribution \(x = (x_1, \ldots, x_n)^{\top}\), the items are ranked in decreasing order of their value in \(x\), with equal values resulting in tied items.
All Markov chain methods take about \(\Theta(n^2N + n^3)\) time \cite{dwork_rank_2001}, although, in practice, it can be reduced to \(\mathcal{O}(n^2N)\) if the transition matrix is explicitly computed.
%
%

\subsubsection*{Maximal lottery}

The \textit{maximal lottery} algorithm is based on a \textit{game-theoretic} model \cite{fishburn_probabilistic_1984}, where the set of items \(\I\) is treated as a set of alternatives.
Let
\begin{equation*}
    \Delta(\I) = \big\{p \in \left[0, 1\right]^{\I} : \sum\nolimits_{x \in \I}  p(x) = 1\big\}
\end{equation*}
be the set of all lotteries over alternatives, where \(p(x)\) is the probability of selecting alternative \(x\).
The pair order matrix \(C\) is converted into a comparison matrix \(M\) by assigning \(M(u, v) = C(u, v)\), for all \(u, v \in \I\) where \(u \neq v\); and setting \(M(u, u) = 0\), for all \(u \in \I\).
This matrix \(M\) then induces a skew-symmetric matrix
\begin{equation*}
    \widetilde{M} = M - M^{T},
\end{equation*}
which can be interpreted as a symmetric zero-sum game.

According to the \textit{minimax} theorem \cite{von_mathematische_1928}, there exists a lottery \(p_{max} = (p_{1}, \ldots, p_{n})^{\top}\) that performs at least as well as any other lottery, meaning:
\begin{equation*}
    \exists p_{max} \in \Delta(\I) : p_{max}^{T}\widetilde{M} \ge 0.
\end{equation*}
This lottery, \(p_{max}\), is referred to as the maximal lottery.
After computing \(p_{max}\), the items are ranked in decreasing order of their values in \(p_{max}\), with ties occurring if the values are equal.

The problem of finding the maximal lottery can be expressed as a \textit{linear programming} task \cite{von_mathematische_1928}, and several strategies have been developed to solve it efficiently \cite{daskalakis_complexity_2009}. %
%
%

\section{Experimental evaluation}
\label{section:experiments}

This section presents an empirical evaluation of the alternative rank aggregation algorithms, focusing on assessing their performance in addressing the partial label ranking problem.

\subsection{Methodology}

The metric of interest is the \(\tau_X\) \textit{rank correlation coefficient} \cite{emond_new_2002}.
Similar to how \textit{Kendall's} \(\tau\) is used for label ranking \cite{cheng_2009_decision}, \(\tau_X\) is specifically designed to deal with tied rankings.
As a result, it is a valuable and widely adopted metric for evaluating partial label ranking performance \cite{alfaro_learning_2021}.
\(\tau_X\) ranges from \(-1\) (complete disagreement) to \(1\) (perfect agreement), with \(0\) indicating no correlation between the predicted and true bucket orders.
Algorithms were assessed using five repetitions of a ten-fold (\(5 \times 10\)) cross-validation method.
To test robustness to missing information, 0\%, 30\%, and 60\% of class labels were randomly omitted.
These percentages are standard values used for label and partial label ranking problems \cite{alfaro_learning_2021, alfaro_ensemble_2023, cheng_label_2010}.
Finally, the results were analyzed following the procedure in \cite{demsar_statistical_2006, garcia_extension_2008} using the \texttt{exreport} tool \cite{arias_exreport:_2015}.
A \textit{Friedman test} \cite{friedman_comparison_1940} was conducted to determine whether the performance differences among algorithms are statistically significant.
This non-parametric test ranks algorithms across datasets and checks if the rank differences are greater than expected by chance.
If the null hypothesis is rejected, a \textit{Holm's post-hoc test} \cite{holm_simple_1979} is applied to compare the top-ranked method against the others, while controlling the family-wise error rate.
The Friedman test and the post-hoc analysis were performed at a 5\% significance level.

\subsection{Datasets and code}

Table \ref{table:datasets} shows the datasets used in the experimental evaluation.
Those above the dashed line are synthetic, generated by transforming multi-class datasets from the UCI repository \cite{kelly_uci} into partial label ranking data following the procedure from \cite{alfaro_learning_2021}.
The datasets below the dashed line represent real-world partial label ranking problems \cite{kelly_uci, maxwell_movielens_2016}.

The table columns are as follows: Identifier, the dataset's unique identifier in the \texttt{OpenML} repository \cite{vanschoren_openml_2014}; \#of instances, the total number of data points; \#of features, the total number of input variables per instance; \#of classes, the total number of class labels; Unique \#of rankings, the number of distinct partial label rankings; and Mean \#of buckets, the average number of buckets per ranking.

\begin{table}[!ht]
    \centering
    \caption{Description of the datasets}
    \label{table:datasets}
    \resizebox{\columnwidth}{!}
    {%
        \begin{tabular}{lrrrrrr}
            \toprule
            Dataset & Identifier & \#of instances & \#of features & \#of classes & Unique \#of rankings & Mean \#of buckets \\
            \midrule
            authorship & 42835 & 841 & 70 & 4 & 47 & 3.063 \\
            blocks & 42836 & 5472 & 10 & 5 & 116 & 2.337 \\
            breast & 42838 & 109 & 9 & 6 & 62 & 3.925 \\
            ecoli & 42844 & 336 & 7 & 8 & 179 & 4.140 \\
            glass & 42848 & 214 & 9 & 6 & 105 & 4.089 \\
            iris  & 42871 & 150 & 4 & 3 & 7 & 2.380 \\
            letter & 42853 & 20000 & 16 & 26 & 15014 & 7.033 \\
            libras & 42855 & 360 & 90 & 15 & 356 & 6.889 \\
            pendigits & 42857 & 10992 & 16 & 10 & 3327 & 3.397 \\
            satimage & 42858 & 6435 & 36 & 6 & 504 & 3.356 \\
            segment & 42860 & 2310 & 18 & 7 & 271 & 3.031 \\
            vehicle & 42864 & 846 & 18 & 4 & 47 & 3.117 \\
            vowel & 42866 & 528 & 10 & 11 & 504 & 5.739 \\
            wine & 42872 & 178 & 13 & 3 & 11 & 2.680 \\
            yeast & 42870 & 1484 & 8 & 10 & 1006 & 5.929 \\
            \hdashline \\[-0.2cm]
            algae & 45755 & 316 & 18 & 7 & 316 & 4.877 \\
            movies & 45738 & 260 & 64 & 15 & 260 & 3.046 \\
            \bottomrule
        \end{tabular}
    }
\end{table}

Note that a higher number of unique rankings relative to the total number of instances indicates greater variability, which typically makes the dataset more challenging to learn.

All code used in the experiments is publicly available at: \url{https://github.com/jiayiwang-lmu/scikit-lr-aggregation}.

\subsection{Learning methods}
\label{section:algorithms}

We used the \textit{partial label ranking trees} algorithm with the \textit{entropy} criterion \cite{alfaro_learning_2021} as the base learning method.
This setup ensures that all aggregation techniques share the same decision tree structure, learned independently of the consensus bucket order.
The differences lie solely in the bucket orders assigned to the leaf nodes during the learning phase, which are determined by the respective aggregation technique.
In other words, the focus here is explicitly on the aggregation step (see Figure \ref{figure:plr}), enabling a targeted investigation into the influence of the chosen aggregation method.
Moreover, this algorithm can predict a set of rankings rather than just a pair order matrix, which makes it possible to evaluate the first three Markov chain-based aggregation methods, as stated in Section \ref{section:markov}.

As explained in Section \ref{section:scoring}, the Borda and Copeland algorithms include a \(\beta\) hyperparameter to manage the items ranked in the same position, specifically, when their scores fall within the overlap range defined by \(\beta\).
As expected, this hyperparameter significantly affects performance.
To examine its impact, we conducted a preliminary experiment to analyze its effect.
In particular, we tested \(\beta\) values ranging from 0 to 2 on the synthetic datasets to identify the value that achieves the best average performance across these datasets.
Note that real-world datasets were excluded, as a different procedure, outlined in Section \ref{section:real}, is used to determine their \(\beta\) values.

\begin{figure}[!ht]
    \centering
    \includegraphics[width = 7.5cm]{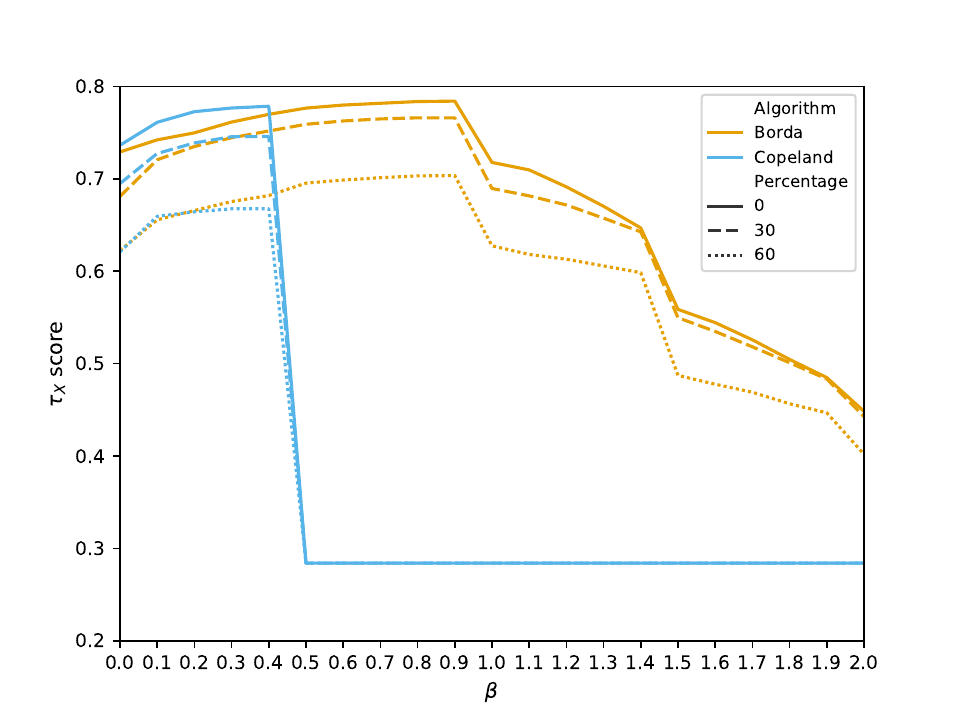}
    \caption{Average \(\tau_X\) score across synthetic datasets for different \(\beta\) values, comparing algorithms under varying percentages of missing class~labels}
    \label{figure:beta}
\end{figure}

Figure \ref{figure:beta} shows the impact of the \(\beta\) hyperparameter on the \(\tau_X\) score for the Borda and Copeland algorithms for varying percentages of missing class~labels.
Notably, the best \(\beta\) value is 0.9 for Borda and 0.4 for Copeland, regardless of the percentage of missing class labels.
As \(\beta\) increases, both methods become more likely to group items into the same bucket, with the extreme case being a single bucket for all items.
For Copeland, this critical point occurs at \(\beta = 0.5\) because the scoring function assigns all items a score of 0.5.
This is because the entire pair order matrix falls within the interval \([0.5 - \beta, 0.5 + \beta] = [0, 1]\), matching its full value range, as discussed in Section \ref{section:obop}.
Therefore, in the rest of the experimental evaluation, we set \(\beta = 0.9\) for Borda and \(\beta = 0.4\) for Copeland.

\subsection{Results}

This section presents the results of our experimental evaluation on both synthetic and real-world datasets.
The aggregation methods from Section \ref{section:aggregations} are referred to by their respective names.
In contrast, the representative aggregation method for the optimal bucket order problem, the current state-of-the-art bucket pivot algorithm (see Section \ref{section:obop}), is denoted as \textit{Bucket pivot} in the upcoming figures.

\subsubsection{Synthetic datasets}
\label{section:synthetic}

Table \ref{table:holm} summarizes the results of the statistical tests.
Bold values indicate algorithms not statistically significantly different from the top-ranked one according to the Friedman test.
The columns are: Rank, average rank from the Friedman test; P-value, adjusted \(p\)-value using the Holm's procedure; and Win, Tie, and Loss, indicating how often the control algorithm wins, ties, or loses to the respective method.
Complete accuracy results for all datasets and missing label settings are provided in the supplementary material.

\begin{table*}[t]
    \caption{Friedman's and Holm's tests with varying percentages of missing class labels}
    \centering
    \setlength{\tabcolsep}{0.15cm}
    \label{table:holm}
    \resizebox{\textwidth}{!}
    {%
        \begin{tabu}{|[0.05cm]lrrrrr|}
            \tabucline[0.05cm]{-}
            \multicolumn{6}{|[0.05cm]c|}{\textbf{Missing percentage: 0\%}} \\
            \hline
            \multicolumn{6}{|[0.05cm]c|}{Friedman p-value: \(\bf 5.431 \times 10^{-18}\)} \\
            \hline
            \multicolumn{6}{|[0.05cm]c|}{Holm results} \\
            \hline
            Method & Rank & P-value & Win & Tie & Loss \\
            \hline
            Bucket pivot & 1.53 & - & - & - & \\
            Borda & 2.03 & \(\bf 5.52 \times 10^{-1}\) & 10 & 1 & 6 \\
            Copeland & 2.68 & \(\bf 3.44 \times 10^{-1}\) & 14 & 1 & 2 \\
            Markov chain 3 & 4.82 & \(2.65 \times 10^{-4}\) & 17 & 0 & 0 \\
            Markov chain 1 & 5.26 & \(3.50 \times 10^{-5}\) & 17 & 0 & 0 \\
            Markov chain 2 & 5.44 & \(1.61 \times 10^{-5}\) & 17 & 0 & 0 \\
            Markov chain 4 & 6.59 & \(1.04 \times 10^{-8}\) & 17 & 0 & 0 \\
            Maximal lottery & 7.65 & \(2.31 \times 10^{-12}\) & 17 & 0 & 0 \\
            \tabucline[0.05cm]{-}
        \end{tabu}%
        \begin{tabu}{lrrrrr}
            \tabucline[0.05cm]{-}
            \multicolumn{6}{c|}{\textbf{Missing percentage: 30\%}} \\
            \hline
            \multicolumn{6}{c|}{Friedman p-value: \(\bf 2.033 \times 10^{-18}\)} \\
            \hline
            \multicolumn{6}{c|}{Holm results} \\
            \hline
            Method & Rank & P-value & Win & Tie & Loss \\
            \hline
            Borda & 1.44 & - & - & - & \\
            Bucket pivot & 2.09 & \(\bf 4.41 \times 10^{-1}\) & 12 & 1 & 4 \\
            Copeland & 2.82 & \(\bf 2.00 \times 10^{-1}\) & 15 & 0 & 2 \\
            Markov chain 3 & 4.38 & \(1.40 \times 10^{-3}\) & 17 & 0 & 0 \\
            Markov chain 2 & 5.09 & \(5.68 \times 10^{-5}\) & 17 & 0 & 0 \\
            Markov chain 1 & 5.71 & \(1.93 \times 10^{-6}\) & 17 & 0 & 0 \\
            Markov chain 4 & 7.06 & \(1.37 \times 10^{-10}\) & 17 & 0 & 0 \\
            Maximal lottery & 7.41 & \(8.34 \times 10^{-10}\) & 16 & 0 & 1 \\
            \tabucline[0.05cm]{-}
        \end{tabu}%
        \begin{tabu}{|lrrrrr|[0.05cm]}
            \tabucline[0.05cm]{-}
            \multicolumn{6}{c|[0.05cm]}{\textbf{Missing percentage: 60\%}} \\
            \hline
            \multicolumn{6}{c|[0.05cm]}{Friedman p-value: \(\bf 1.043 \times 10^{-13}\)} \\
            \hline
            \multicolumn{6}{c|[0.05cm]}{Holm results} \\
            \hline
            Method & Rank & P-value & Win & Tie & Loss \\
            \hline
            Borda & 1.35 & - & - & - & \\
            Copeland & 2.88 & \(\bf 1.37 \times 10^{-1}\) & 15 & 0 & 2 \\
            Bucket pivot & 2.88 & \(\bf 1.35 \times 10^{-1}\) & 14 & 0 & 3 \\
            Markov chain 3 & 4.26 & \(1.59 \times 10^{-3}\) & 17 & 0 & 0 \\
            Markov chain 2 & 5.76 & \(6.25 \times 10^{-7}\) & 17 & 0 & 0 \\
            Markov chain 1 & 5.79 & \(6.25 \times 10^{-7}\) & 17 & 0 & 0 \\
            Maximal lottery & 6.47 & \(6.72 \times 10^{-9}\) & 16 & 0 & 1 \\
            Markov chain 4 & 6.59 & \(3.24 \times 10^{-9}\) & 17 & 0 & 0 \\
            \tabucline[0.05cm]{-}
        \end{tabu}%
    }
\end{table*}

\begin{figure}[!ht]
    \centering
    \includegraphics[width = 7.5cm]{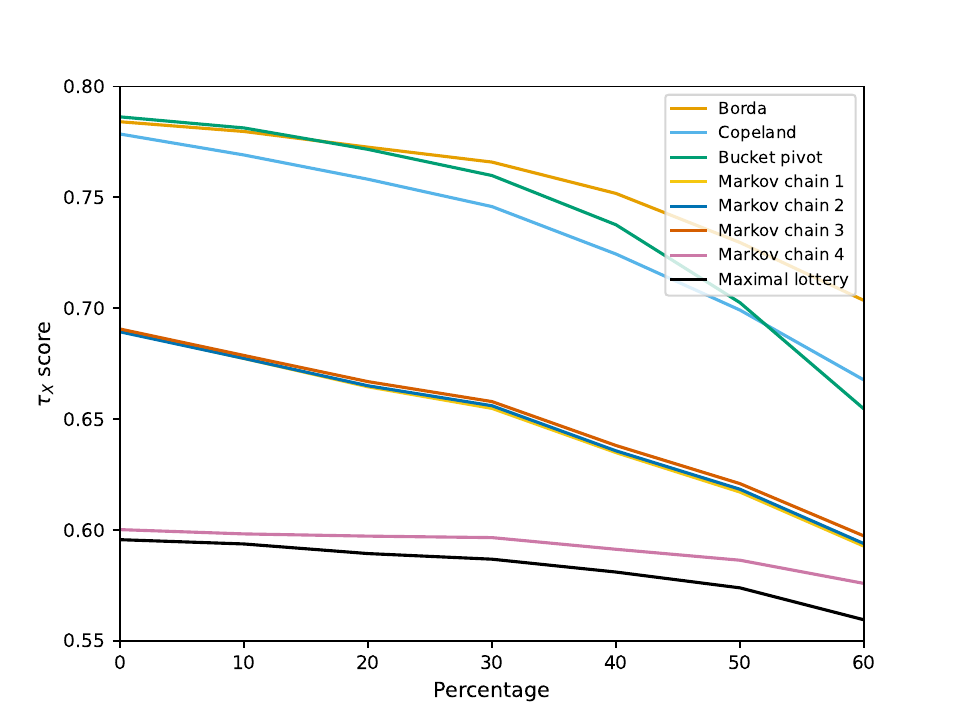}
    \caption{Average \(\tau_X\) score across synthetic datasets for varying percentages of missing class labels, comparing algorithms}
    \label{figure:accuracy}
\end{figure}

This analysis shows that scoring-based methods perform remarkably well across all scenarios.
With incomplete rankings, the Borda algorithm outperforms the bucket pivot method at 30\% of class labels are missing.
As missingness increases, both Borda and Copeland demonstrate greater robustness compared to other methods, consistently ranking ahead of the bucket pivot method.
This trend is particularly evident with 60\% of missing data, where Borda and Copeland exhibit significantly better performance, as shown in Figure \ref{figure:accuracy}.
In contrast, probabilistic-based methods perform substantially worse in all settings, consistently ranking below the scoring-based algorithms.

\begin{figure}[!ht]
    \centering
    \includegraphics[width = 7.5cm]{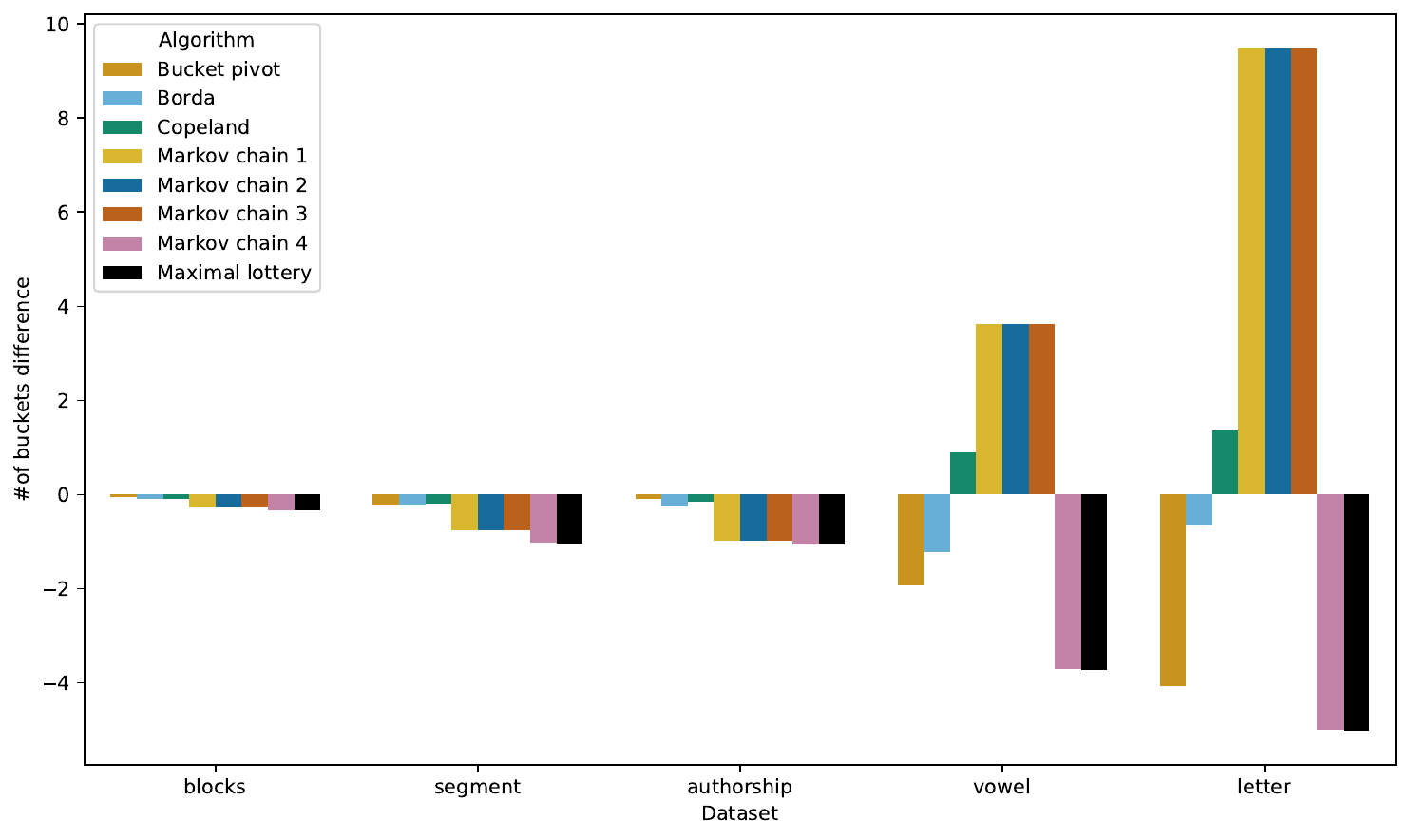}
    \caption{Average bucket count difference between the true and predicted bucket orders across algorithms and a subset of datasets, based on complete rankings.}
    \label{figure:buckets}
\end{figure}

The accuracy metric (i.e., the \(\tau_X\) rank correlation coefficient) penalizes two specific cases: %
\begin{enumerate*}[label = (\roman*)]%
    \item when items ranked ahead or tied in the true ranking appear behind in the prediction, and %
    \item when lower-ranked items in the true ranking are tied in the prediction. %
\end{enumerate*} %
These penalties arise when the structure of the predicted bucket order misrepresents the true pairwise preferences between items.
We hypothesize that such errors result from algorithms producing either too many or too few buckets.
Too many buckets lead to overly fine-grained rankings, forcing distinctions where the ground truth contains ties, resulting in inversions (case i).
Conversely, too few buckets cause distinct items to be grouped together, missing necessary orderings, resulting in false ties (case ii).
Both scenarios reduce the \(\tau_X\) accuracy by misrepresenting the true bucket order structure.
To investigate this, we measured the discrepancy between the number of buckets in the predicted and true bucket orders, as illustrated in Figure \ref{figure:buckets}, where the x-axis is sorted by each dataset's mean number of buckets.
This figure highlights the most notable trends for complete rankings; full results across all datasets and settings are provided in the supplementary material.
Notably, probabilistic-based methods exhibit more pronounced deviations compared to scoring-based algorithms.
Specifically, Markov chain 1, 2, and 3 methods produce overly fine-grained bucket orders, approximating total orders, especially in datasets with many class labels (e.g., vowel or letter).
Conversely, Markov chain 4 and the maximal lottery algorithms tend to produce too few buckets, typically forming one dominant top bucket and grouping remaining items into coarser sets.
In other scenarios, all probabilistic-based methods tend to produce bucket orders with fewer buckets overall (e.g., blocks, segment, or authorship).
Similar patterns were observed in the case of incomplete rankings, as shown in the supplementary material.

Regarding computational complexity, training times (in seconds) for complete rankings are shown in Table \ref{table:time}.
Only the scoring-based algorithms and the bucket pivot method are included, since the probabilistic-based methods were both less accurate and significantly more computationally demanding.
Additionally, prediction times are omitted because they are identical across all algorithms.
This is because the underlying decision tree structure is the same for all the methods, as explained in Section \ref{section:algorithms}.
Overall, Borda and Copeland are about twice as fast as the bucket pivot algorithm, despite the latter having a lower computational complexity, as stated in Sections \ref{section:obop} and \ref{section:scoring}.
This discrepancy likely results from the recursive nature of the bucket pivot method, which adds overhead, thereby increasing its runtime.
Similar timing patterns were observed in the case of incomplete rankings. Full training time results for all settings are provided in the supplementary material.

\begin{table}[t]
    \centering
    \caption{Average training times (mean and standard deviation, in seconds) for scoring-based algorithms (Borda and Copeland) and the bucket pivot method across synthetic datasets under complete rankings.}
    \label{table:time}
    \resizebox{\columnwidth}{!}
    {%
        \begin{tabular}{lrrr}
            \toprule
            Dataset & Bucket pivot & Borda & Copeland \\
            \midrule
            authorship & 0.678 $ \pm $ 0.024 & 0.341 $ \pm $ 0.013 & 0.341 $ \pm $ 0.013 \\
            blocks & 1.232 $ \pm $ 0.050 & 0.697 $ \pm $ 0.034 & 0.695 $ \pm $ 0.031 \\
            breast & 0.027 $ \pm $ 0.003 & 0.014 $ \pm $ 0.002 & 0.014 $ \pm $ 0.002 \\
            ecoli & 0.045 $ \pm $ 0.004 & 0.027 $ \pm $ 0.002 & 0.028 $ \pm $ 0.003 \\
            glass & 0.051 $ \pm $ 0.004 & 0.024 $ \pm $ 0.003 & 0.024 $ \pm $ 0.002 \\
            iris & 0.007 $ \pm $ 0.002 & 0.005 $ \pm $ 0.002 & 0.005 $ \pm $ 0.002 \\
            letter & 18.073 $ \pm $ 0.119 & 17.501 $ \pm $ 0.134 & 17.554 $ \pm $ 0.137 \\
            libras & 2.121 $ \pm $ 0.063 & 1.640 $ \pm $ 0.047 & 1.685 $ \pm $ 0.044 \\
            pendigits & 4.848 $ \pm $ 0.060 & 3.433 $ \pm $ 0.044 & 3.434 $ \pm $ 0.049 \\
            satimage & 3.527 $ \pm $ 0.063 & 2.050 $ \pm $ 0.039 & 2.053 $ \pm $ 0.039 \\
            segment & 1.300 $ \pm $ 0.033 & 0.682 $ \pm $ 0.017 & 0.689 $ \pm $ 0.021 \\
            vehicle & 0.228 $ \pm $ 0.013 & 0.118 $ \pm $ 0.009 & 0.122 $ \pm $ 0.009 \\
            vowel & 0.338 $ \pm $ 0.012 & 0.230 $ \pm $ 0.009 & 0.234 $ \pm $ 0.011 \\
            wine & 0.041 $ \pm $ 0.004 & 0.018 $ \pm $ 0.002 & 0.020 $ \pm $ 0.006 \\
            yeast & 0.256 $ \pm $ 0.011 & 0.161 $ \pm $ 0.010 & 0.162 $ \pm $ 0.008 \\
            \bottomrule
        \end{tabular}
    }
\end{table}

\subsubsection{Real-world datasets}
\label{section:real}

The preliminary analysis of the \(\beta\) hyperparameter in Section \ref{section:algorithms}  showed its importance for the performance of scoring-based algorithms.
While a fixed value may suffice across all datasets, it is compelling to establish a criterion for selecting suitable values in real-world settings.
This section introduces a recommendation system based on the experimental results from synthetic datasets.

We trained a linear regression model using the following dataset meta-features as predictors for the new datasets: number of instances, number of features, number of classes, average number of buckets, and average percentage of missing class labels.
Thus, each training sample represents a unique combination of a dataset and a specific percentage of missing class labels, with percentages ranging from 0 to 60 in increments of 10.
These averages were computed from the \(5 \times 10\) cross-validation method carried out.
The target variable for the model was the \(\beta\) yielding the best average results.

Table \ref{table:real} shows the results obtained from the cross-validation study, where \(^*\) indicates that the results correspond to the best identified for the respective algorithm during the cross-validation.
In contrast, Table \ref{table:beta} compares the best \(\beta\) values with those predicted by the linear regression model.
For the algae dataset, the Borda algorithm outperforms the bucket pivot method and achieves results similar to Borda\(^*\), as the predicted \(\beta\) value is very close to the best one.
In the case of the Copeland algorithm, although it does not surpass the bucket pivot method, the predicted \(\beta\) value remains close to the best one.
On the other hand, the probabilistic-based algorithms fail to produce competitive results.
For the movies dataset, neither Borda nor Copeland delivers competitive results, as their predicted \(\beta\) values are not close to their respective best values.
Interestingly, the maximal lottery algorithm performs better than Borda and Copeland, yet it still fails to achieve the performance achieved by bucket pivot.

\begin{table}[t]
    \centering
    \caption{Average \(\tau_X\) score (mean and standard deviation) across real-world datasets for varying percentages of missing class labels, comparing algorithms}
    \label{table:real}
    \resizebox{\columnwidth}{!}
    {
        \begin{tabular}{lrrrrrr}
            \toprule
            \multirow{2}{*}{Algorithm} & \multicolumn{3}{r}{algae} & \multicolumn{3}{r}{movies} \\
            \cmidrule{2-7}
            & 0\% & 30\% & 60\% & 0\% & 30\% & 60\% \\
            \midrule
            Bucket pivot & 0.419 $ \pm $ 0.064 & 0.381 $ \pm $ 0.049 & 0.285 $ \pm $ 0.052 & 0.439 $ \pm $ 0.032 & 0.433 $ \pm $ 0.032 & 0.433 $ \pm $ 0.029 \\
            Borda & 0.435 $ \pm $ 0.064 & 0.389 $ \pm $ 0.057 & 0.326 $ \pm $ 0.055 & 0.401 $ \pm $ 0.033 & 0.380 $ \pm $ 0.030 & 0.373 $ \pm $ 0.039 \\
            Borda\(^*\) & 0.437 $ \pm $ 0.064 & 0.395 $ \pm $ 0.059 & 0.330 $ \pm $ 0.054 & 0.437 $ \pm $ 0.027 & 0.440 $ \pm $ 0.028 & 0.442 $ \pm $ 0.025 \\
            Copeland & 0.414 $ \pm $ 0.065 & 0.362 $ \pm $ 0.055 & 0.283 $ \pm $ 0.052 & 0.265 $ \pm $ 0.038 & 0.192 $ \pm $ 0.032 & 0.180 $ \pm $ 0.037 \\
            Copeland\(^*\) & 0.420 $ \pm $ 0.065 & 0.361 $ \pm $ 0.056 & 0.286 $ \pm $ 0.053 & 0.443 $ \pm $ 0.028 & 0.443 $ \pm $ 0.028 & 0.443 $ \pm $ 0.028 \\
            Markov chain 1 & 0.390 $ \pm $ 0.060 & 0.344 $ \pm $ 0.058 & 0.294 $ \pm $ 0.065 & 0.179 $ \pm $ 0.036 & 0.135 $ \pm $ 0.032 & 0.126 $ \pm $ 0.036 \\
            Markov chain 2 & 0.384 $ \pm $ 0.059 & 0.348 $ \pm $ 0.056 & 0.293 $ \pm $ 0.069 & 0.180 $ \pm $ 0.036 & 0.145 $ \pm $ 0.032 & 0.136 $ \pm $ 0.038 \\
            Markov chain 3 & 0.394 $ \pm $ 0.058 & 0.355 $ \pm $ 0.057 & 0.300 $ \pm $ 0.067 & 0.179 $ \pm $ 0.036 & 0.147 $ \pm $ 0.032 & 0.138 $ \pm $ 0.038 \\
            Markov chain 4 & 0.364 $ \pm $ 0.057 & 0.339 $ \pm $ 0.064 & 0.275 $ \pm $ 0.053 & 0.385 $ \pm $ 0.028 & 0.250 $ \pm $ 0.041 & 0.165 $ \pm $ 0.039 \\
            Maximal lottery & 0.348 $ \pm $ 0.055 & 0.328 $ \pm $ 0.055 & 0.299 $ \pm $ 0.057 & 0.411 $ \pm $ 0.025 & 0.398 $ \pm $ 0.025 & 0.377 $ \pm $ 0.033 \\
            \bottomrule
        \end{tabular}
    }
\end{table}

\begin{table}[t]
    \centering
    \caption{Comparison of the best and predicted \(\beta\) values for scoring-based algorithms across real-world datasets and percentages of missing class labels}
    \label{table:beta}
    \resizebox{\columnwidth}{!}
    {%
        \begin{tabular}{llrrrrrr}
            \toprule
            \multicolumn{2}{c}{\multirow{2}{*}{Algorithm}} & \multicolumn{3}{r}{algae} & \multicolumn{3}{r}{movies} \\
            \cmidrule{3-8}
            & & 0\% & 30\% & 60\% & 0\% & 30\% & 60\% \\
            \midrule
            \multirow{2}{*}{Borda} & Best & 0.8 & 0.8 & 0.8 & 1.9 & 2.0 & 2.0 \\
            & Predicted & 0.677 $ \pm $ 0.002 & 0.652 $ \pm $ 0.002 & 0.634 $ \pm $ 0.002 & 1.145 $ \pm $ 0.001 & 1.063 $ \pm $ 0.003 & 0.993 $ \pm $ 0.003 \\
            \midrule
            \multirow{2}{*}{Copeland} & Best & 0.2 & 0.3 & 0.3 & 0.5 & 0.5 & 0.5 \\
            & Predicted & 0.303 $ \pm $ 0.00 & 0.266 $ \pm $ 0.001 & 0.229 $ \pm $ 0.001 & 0.273 $ \pm $ 0.000 & 0.226 $ \pm $ 0.001 & 0.181 $ \pm $ 0.001 \\
            \bottomrule
        \end{tabular}
    }
\end{table}

\section{Conclusions}
\label{section:conclusions}

This paper has explored alternative rank aggregation methods for the partial label ranking problem.
Specifically, we investigated two popular classes, scoring-based and non-parametric probabilistic-based, and modified the former to reduce its tendency to produce overly stringent total orders.
Our experiments showed that scoring-based methods are competitive with the current state-of-the-art algorithm when using complete rankings and outperform it with incomplete rankings.
Conversely, non-parametric probabilistic-based algorithms failed to achieve competitive performance. 
Regarding computational efficiency, scoring-based algorithms are generally twice as fast as the current state-of-the-art approach, primarily due to the latter's recursive nature.
For future work, we aim to enhance probabilistic-based methods, potentially considering also parametric variants, and extend them to support ties. Moreover, we plan to improve the recommendation model for the \(\beta\) hyperparameter by incorporating more useful meta-characteristics from the datasets.

\begin{ack}

    This work is partially funded by the following projects: SBPLY/21/180225/000062 (Junta de Comunidades de Castilla-La Mancha and ERDF A way of making Europe) and PID2022-139293NB-C32 (MICIU/AEI/10.13039/501100011033 and ERDF, EU).

\end{ack}


\appendix
\onecolumn
\section{Aggregation algorithms}

This section provides the pseudocode of the aggregation algorithms evaluated in the main study.


\begin{algorithm}
    \caption{Copeland}
    \label{algorithm:copeland}
    \begin{algorithmic}[1]
        \REQUIRE Set of items $ \mathcal{I} $; Pair order matrix $ C $; Hyperparameter $ \beta \geq 0 $
        \ENSURE Consensus bucket order $ \mathcal{B} $
        \vspace{0.15cm}
        \hrule
        \vspace{0.05cm}
        \STATE Initialize scores $ S(u) \gets 0 $ for all \(u \in \I\)
        \FOR{$ u \in \mathcal{I} $}
            \FOR{$ v \in \mathcal{I} $}
                \IF{$ u \neq v $}
                    \IF{$ C(u, v) > 0.5 + \beta $}
                        \STATE $ S(u) \gets S(u) + 1 $
                    \ELSIF{$ 0.5 - \beta \leq C(u, v) \leq 0.5 + \beta $}
                        \STATE $ S(u) \gets S(u) + 0.5 $
                    \ENDIF
                \ENDIF
            \ENDFOR
        \ENDFOR
        \STATE Compute bucket order $ \mathcal{B} $ according to the rules:
        \begin{align*}
            u \succ v & \quad \text{if } S(u) > S(v), \\
            u \sim v & \quad \text{if } S(u) = S(v), \\
            u \prec v & \quad \text{if } S(u) < S(v).
        \end{align*}
        \RETURN $ \mathcal{B} $
    \end{algorithmic}
\end{algorithm}


\begin{algorithm}
    \caption{Borda}
    \label{algorithm:borda}
    \begin{algorithmic}[1]
        \REQUIRE Set of items $ \mathcal{I} $; Pair order matrix $ C $; Hyperparameter $ \beta \geq 0 $
        \ENSURE Consensus bucket order $ \mathcal{B} $
        \vspace{0.15cm}
        \hrule
        \vspace{0.05cm}
        \STATE Initialize scores $ S(u) \gets 0 $ for all \(u \in \I\)
        \FOR{$ u \in \mathcal{I} $}
            \FOR{$ v \in \mathcal{I} $}
                \IF{$ u \neq v $}
                    \STATE $ S(u) \gets S(u) + C(u, v) $
                \ENDIF
            \ENDFOR
        \ENDFOR
        \STATE Compute bucket order $ \mathcal{B} $ according to the rules:
        \begin{align*}
            u \succ v & \text{ if } S(u) - S(v) > \beta, \\
            u \sim v & \text{ if } S(u) - S(v) \in [-\beta, \beta] \\
            u \prec v & \text{ if } S(u) - S(v) < -\beta.
         \end{align*}
        \RETURN $ \mathcal{B} $
    \end{algorithmic}
\end{algorithm}


\begin{algorithm}
    \caption{Markov chain 1}
    \label{algorithm:mc1}
    \begin{algorithmic}[1]
        \REQUIRE Set of items $ \mathcal{I} $; Set of (possibly incomplete) partial rankings $ \mathbf{\Pi} = \{\pi_1, \dots, \pi_N\} $        
        \ENSURE Consensus bucket order $ \mathcal{B} $
        \vspace{0.15cm}
        \hrule
        \vspace{0.05cm}
        \STATE Initialize transition matrix $ P \gets 0 $
        \FOR{$ u \in \mathcal{I} $}
            \FOR{$ v \in \mathcal{I} $}
                \STATE $ y_{uv} \gets \left|\left\{i : \pi_i(v) \leq \pi_i(u)\right\}_{i = 1}^{N}\right| $ \COMMENT{Number of rankings where $ v $ is ranked higher or equal to $ u $}
                \STATE $ z_u \gets \sum_{i = 1}^{N} \left|\left\{w : \pi_i(w) \leq \pi_i(u) \right\}\right| $ \COMMENT{Number of items ranked higher or equal to $ u $ in each ranking $ \pi_i $}
                \STATE $ P(uv) \gets \frac{y_{uv}}{z_u} $
            \ENDFOR
        \ENDFOR
    \STATE Find stationary distribution that satisfies $ x = xP $
    \STATE Compute bucket order $ \mathcal{B} $ based on $ x $:
    \begin{align*}
        u \succ v & \text{ if } x_u > x_v \\
        u \sim v & \text{ if } x_u = x_v \\
        u \prec v & \text{ if } x_u < x_v
    \end{align*}
    \RETURN $ \mathcal{B} $
    \end{algorithmic}
\end{algorithm}


\begin{algorithm}
    \caption{Markov chain 2}
    \label{algorithm:mc2}
    \begin{algorithmic}[1]
        \REQUIRE Set of items $ \mathcal{I} $; Set of (possibly incomplete) partial rankings $ \mathbf{\Pi} = \{\pi_1, \dots, \pi_N\} $        
        \ENSURE Consensus bucket order $ \mathcal{B} $
        \vspace{0.15cm}
        \hrule
        \vspace{0.05cm}
        \STATE Initialize transition matrix $ P \gets 0 $
        \FOR{$ u \in \mathcal{I} $}
            \FOR{$ v \in \mathcal{I} $}
                \FOR{$ \pi_i \in \Pi $}
                    \STATE $ y_{ui} = \left|\left\{w : \pi_i(w) \leq \pi_i(u) \right\}\right| $ \COMMENT{Number of items ranked higher or equal to $ u $ in $ \pi_i $}
                    \STATE $ P(u, v) \gets \frac{\mathbf{1}_{\left\{\pi_i(v) \leq \pi_i(u)\right\}}}{y_{ui}} $
                \ENDFOR
            \ENDFOR
        \ENDFOR
    \STATE Find stationary distribution that satisfies $ x = xP $
    \STATE Compute bucket order $ \mathcal{B} $ based on $ x $:
    \begin{align*}
        u \succ v & \text{ if } x_u > x_v \\
        u \sim v & \text{ if } x_u = x_v \\
        u \prec v & \text{ if } x_u < x_v
    \end{align*}
    \RETURN $ \mathcal{B} $
    \end{algorithmic}
\end{algorithm}


\begin{algorithm}
    \caption{Markov chain 3}
    \label{algorithm:mc3}
    \begin{algorithmic}[1]
        \REQUIRE Set of items $ \mathcal{I} $; Set of (possibly incomplete) partial rankings $ \mathbf{\Pi} = \{\pi_1, \dots, \pi_N\} $        
        \ENSURE Consensus bucket order $ \mathcal{B} $
        \vspace{0.15cm}
        \hrule
        \vspace{0.05cm}
        \STATE Initialize transition matrix $ P \gets 0 $
        \FOR{$ u \in \mathcal{I} $}
            \FOR{$ v \in \mathcal{I} $}
                \IF{$ u = v $}
                    \STATE $ y_{uv} \gets \sum_{i = 1}^{N} n - \left|\left\{w : \pi_i(w) \leq \pi_i(u) \right\}\right|  + 1 $
                \ELSE
                    \STATE
                    \STATE $ y_{uv} \gets \left|\left\{i : \pi_i(v) \leq \pi_i(u)\right\}_{i = 1}^{N}\right| $ \COMMENT{Number of rankings where $ v $ is ranked higher or equal to $ u $}
                \ENDIF
                \STATE $ P(u, v) \gets \frac{y_uv}{kn} $
            \ENDFOR
        \ENDFOR
    \STATE Find stationary distribution that satisfies $ x = xP $
    \STATE Compute bucket order $ \mathcal{B} $ based on $ x $:
    \begin{align*}
        u \succ v & \text{ if } x_u > x_v \\
        u \sim v & \text{ if } x_u = x_v \\
        u \prec v & \text{ if } x_u < x_v
    \end{align*}
    \RETURN $ \mathcal{B} $
    \end{algorithmic}
\end{algorithm}


\begin{algorithm}
    \caption{Markov chain 4}
    \label{algorithm:mc4}
    \begin{algorithmic}[1]
        \REQUIRE Set of items $ \mathcal{I} $; Pair order matrix $ C $
        \ENSURE Consensus bucket order $ \mathcal{B} $
        \vspace{0.15cm}
        \hrule
        \vspace{0.05cm}
        \STATE Initialize the transition matrix $ P \gets 0 $
        \FOR{$ u \in \mathcal{I} $}
            \FOR{$ v \in \mathcal{I} $}
                \IF {$ u \neq v $}
                    \IF{$ C(u, v) \leq 0.5 $}
                        \STATE $ P(u, v) = \frac{1}{n} $
                    \ELSIF{$ C(u, v) > 0.5 $}
                        \STATE $ P(u, v) = 0 $
                    \ENDIF
                \ENDIF
            \ENDFOR
        \ENDFOR
        \FOR{$ u \in \mathcal{I} $}
            \STATE $ P(u, u) \gets 1 - \sum_{v \in \I:v \neq u} P(u, v) $
        \ENDFOR    
        \STATE Find stationary distribution that satisfies $ x = xP $
        \STATE Compute bucket order $ \mathcal{B} $ based on $ x $:
        \begin{align*}
            u \succ v & \text{ if } x_u > x_v \\
            u \sim v & \text{ if } x_u = x_v \\
            u \prec v & \text{ if } x_u < x_v
        \end{align*}
        \RETURN $ \mathcal{B} $
    \end{algorithmic}
\end{algorithm}


\begin{algorithm}
    \caption{Maximal lottery}
    \label{algorithm:ml}
    \begin{algorithmic}[1]
        \REQUIRE Set of items $ \mathcal{I} $; Pair order matrix $ C $
        \ENSURE Consensus bucket order $ \mathcal{B} $
        \vspace{0.15cm}
        \hrule
        \vspace{0.05cm}
        \STATE Initialize comparison matrix $ M \gets C $
        \STATE $ M(u, u) \gets 0 $ for all \(u \in \I\)
        \STATE Compute the skew-symmetric matrix $ \widetilde{M} \gets M - M^T $
        \STATE Find maximal lottery $ p_{max} $ that satisfies $ p_{max}^T \widetilde{M} \geq 0 $
        \STATE Compute bucket order $ \mathcal{B} $ based on $ p_{max} $:
        \begin{align*}
            u \succ v &\text{ if } p_u > p_v \\
            u \sim v &\text{ if } p_u = p_v \\
            u \prec v &\text{ if } p_u < p_v
        \end{align*}
        \RETURN $ \mathcal{B} $
    \end{algorithmic}
\end{algorithm}

\clearpage
\newpage

\section{Complete accuracy results}

This section presents the complete accuracy results of all aggregation algorithms evaluated across 16 synthetic datasets under varying percentage of missing class labels: 0\%, 30\%, and 60\%. The performance metric used is the average \(\tau_X\) score, reported with its standard deviation over multiple runs. In these tables, the \underline{underlined} values indicate the algorithm(s) that achieved the highest \(\tau_X\) score. \textbf{Boldfaced} values, meanwhile, denote algorithms whose 95\% confidence intervals, defined by their mean and standard deviation, overlap with the interval of the best-performing algorithm.

\begin{table}[ht]
    \centering
    \caption{Average \(\tau_X\) score (mean and standard deviation) of algorithms with 0\% of missing class label positions for the synthetic datasets}
    \label{table:accuracy_0}
    \resizebox{\textwidth}{!}
    {%
        \begin{tabular}{lrrrrrrrr}
            \toprule
            Dataset & Bucket pivot & Borda & Copeland & Markov chain 1 & Markov chain 2 & Markov chain 3 & Markov chain 4 & Maximal lottery \\
            \midrule
            authorship & \underline{$ \bf 0.780 \pm 0.024 $} & $ \bf 0.779 \pm 0.022 $ & $ \bf 0.776 \pm 0.021 $ & $ 0.670 \pm 0.024 $ & $ 0.670 \pm 0.024 $ & $ 0.670 \pm 0.024 $ & $ 0.659 \pm 0.022 $ & $ 0.657 \pm 0.023 $ \\
            blocks & \underline{$ \bf 0.944 \pm 0.004 $} & $ \bf 0.941 \pm 0.005 $ & $ \bf 0.943 \pm 0.004 $ & $ 0.915 \pm 0.006 $ & $ 0.915 \pm 0.006 $ & $ 0.915 \pm 0.006 $ & $ 0.904 \pm 0.007 $ & $ 0.903 \pm 0.007 $ \\
            breast & \underline{$ \bf 0.769 \pm 0.059 $} & \underline{$ \bf 0.769 \pm 0.055 $} & $ \bf 0.762 \pm 0.060 $ & $ \bf 0.693 \pm 0.071 $ & $ \bf 0.693 \pm 0.070 $ & $ \bf 0.695 \pm 0.069 $ & $ 0.507 \pm 0.054 $ & $ 0.504 \pm 0.055 $ \\
            ecoli & \underline{$ \bf 0.763 \pm 0.033 $} & $ \bf 0.756 \pm 0.033 $ & $ \bf 0.739 \pm 0.030 $ & $ 0.632 \pm 0.026 $ & $ 0.631 \pm 0.027 $ & $ 0.632 \pm 0.026 $ & $ 0.558 \pm 0.035 $ & $ 0.551 \pm 0.034 $ \\
            glass & \underline{$ \bf 0.760 \pm 0.036 $} & $ \bf 0.747 \pm 0.036 $ & \underline{$ \bf 0.760 \pm 0.038 $} & $ \bf 0.615 \pm 0.038 $ & $ 0.615 \pm 0.037 $ & $ 0.614 \pm 0.037 $ & $ 0.485 \pm 0.031 $ & $ 0.480 \pm 0.031 $ \\
            iris & $ \bf 0.915 \pm 0.043 $ & $ \bf 0.918 \pm 0.046 $ & \underline{$ \bf 0.919 \pm 0.045 $} & $ \bf 0.854 \pm 0.059 $ & $ \bf 0.854 \pm 0.059 $ & $ \bf 0.854 \pm 0.059 $ & $ \bf 0.848 \pm 0.064 $ & $ \bf 0.845 \pm 0.061 $ \\
            letter & \underline{$ \bf 0.669 \pm 0.005 $} & $ \bf 0.663 \pm 0.004 $ & $ \bf 0.661 \pm 0.005 $ & $ 0.601 \pm 0.006 $ & $ 0.600 \pm 0.006 $ & $ 0.603 \pm 0.006 $ & $ 0.525 \pm 0.007 $ & $ 0.523 \pm 0.007 $ \\
            libras & $ \bf 0.575 \pm 0.026 $ & \underline{$ \bf 0.601 \pm 0.025 $} & $ \bf 0.571 \pm 0.026 $ & $ \bf 0.552 \pm 0.024 $ & $ \bf 0.548 \pm 0.024 $ & $ \bf 0.555 \pm 0.024 $ & $ 0.317 \pm 0.026 $ & $ 0.307 \pm 0.024 $ \\
            pendigits & \underline{$ \bf 0.813 \pm 0.006 $} & $ \bf 0.797 \pm 0.005 $ & $ \bf 0.790 \pm 0.006 $ & $ 0.740 \pm 0.006 $ & $ 0.740 \pm 0.006 $ & $ 0.741 \pm 0.006 $ & $ 0.717 \pm 0.007 $ & $ 0.716 \pm 0.007 $ \\
            satimage & \underline{$ \bf 0.846 \pm 0.006 $} & $ \bf 0.841 \pm 0.005 $ & $ \bf 0.837 \pm 0.007 $ & $ 0.707 \pm 0.009 $ & $ 0.707 \pm 0.009 $ & $ 0.707 \pm 0.009 $ & $ 0.659 \pm 0.009 $ & $ 0.656 \pm 0.009 $ \\
            segment & \underline{$ \bf 0.894 \pm 0.009 $} & $ \bf 0.892 \pm 0.009 $ & $ \bf 0.889 \pm 0.009 $ & $ 0.792 \pm 0.013 $ & $ 0.792 \pm 0.013 $ & $ 0.792 \pm 0.013 $ & $ 0.764 \pm 0.013 $ & $ 0.762 \pm 0.013 $ \\
            vehicle & \underline{$ \bf 0.791 \pm 0.020 $} & $ \bf 0.788 \pm 0.021 $ & $ \bf 0.790 \pm 0.021 $ & $ 0.656 \pm 0.025 $ & $ 0.656 \pm 0.025 $ & $ 0.656 \pm 0.026 $ & $ 0.582 \pm 0.022 $ & $ 0.572 \pm 0.023 $ \\
            vowel & $ \bf 0.679 \pm 0.023 $ & \underline{$ \bf 0.682 \pm 0.024 $} & $ \bf 0.661 \pm 0.025 $ & $ \bf 0.591 \pm 0.028 $ & $ \bf 0.591 \pm 0.027 $ & $ \bf 0.595 \pm 0.028 $ & $ 0.368 \pm 0.024 $ & $ 0.358 \pm 0.021 $ \\
            wine & \underline{$ \bf 0.821 \pm 0.049 $} & $ \bf 0.815 \pm 0.051 $ & $ \bf 0.815 \pm 0.050 $ & $ \bf 0.737 \pm 0.049 $ & $ \bf 0.737 \pm 0.051 $ & $ \bf 0.737 \pm 0.049 $ & $ \bf 0.739 \pm 0.048 $ & $ \bf 0.734 \pm 0.051 $ \\
            yeast & \underline{$ \bf 0.775 \pm 0.010 $} & $ \bf 0.772 \pm 0.009 $ & $ \bf 0.765 \pm 0.010 $ & $ 0.591 \pm 0.018 $ & $ 0.591 \pm 0.018 $ & $ 0.593 \pm 0.019 $ & $ 0.370 \pm 0.009 $ & $ 0.366 \pm 0.010 $ \\
                        \bottomrule
        \end{tabular}
    }
\end{table}

\begin{table}[ht]
    \centering
    \caption{Average \(\tau_X\) score (mean and standard deviation) of algorithms with 30\% of missing class label positions for the synthetic datasets}
    \label{table:accuracy_30}
    \resizebox{\textwidth}{!}
    {
        \begin{tabular}{lrrrrrrrr}
            \toprule
            Dataset & Bucket pivot & Borda & Copeland & Markov chain 1 & Markov chain 2 & Markov chain 3 & Markov chain 4 & Maximal lottery \\
            \midrule
            authorship & $ \bf 0.733 \pm 0.031 $ & \underline{$ \bf 0.741 \pm 0.028 $} & $ \bf 0.737 \pm 0.028 $ & $ \bf 0.640 \pm 0.026 $ & $ \bf 0.639 \pm 0.026 $ & $ \bf 0.640 \pm 0.026 $ & $ \bf 0.632 \pm 0.028 $ & $ \bf 0.634 \pm 0.030 $ \\
            blocks & $ \bf 0.928 \pm 0.007 $ & \underline{$ \bf 0.933 \pm 0.006 $} & $ \bf 0.926 \pm 0.007 $ & $ 0.906 \pm 0.007 $ & $ 0.906 \pm 0.007 $ & $ 0.906 \pm 0.007 $ & $ 0.896 \pm 0.007 $ & $ 0.901 \pm 0.007 $ \\
            breast & $ \bf 0.715 \pm 0.074 $ & \underline{$ \bf 0.727 \pm 0.064 $} & $ \bf 0.702 \pm 0.060 $ & $ \bf 0.591 \pm 0.081 $ & $ \bf 0.593 \pm 0.076 $ & $ \bf 0.594 \pm 0.076 $ & $ \bf 0.503 \pm 0.061 $ & $ \bf 0.475 \pm 0.065 $ \\
            ecoli & $ \bf 0.730 \pm 0.032 $ & \underline{$ \bf 0.738 \pm 0.031 $} & $ \bf 0.698 \pm 0.037 $ & $ \bf 0.613 \pm 0.040 $ & $ \bf 0.613 \pm 0.040 $ & $ \bf 0.616 \pm 0.038 $ & $ 0.556 \pm 0.038 $ & $ 0.548 \pm 0.036 $ \\
            glass & $ \bf 0.714 \pm 0.046 $ & \underline{$ \bf 0.730 \pm 0.040 $} & $ \bf 0.727 \pm 0.042 $ & $ 0.551 \pm 0.043 $ & $ 0.553 \pm 0.042 $ & $ 0.556 \pm 0.042 $ & $ 0.484 \pm 0.041 $ & $ 0.460 \pm 0.037 $ \\
            iris & \underline{$ \bf 0.903 \pm 0.046 $} & $ \bf 0.896 \pm 0.047 $ & $ \bf 0.902 \pm 0.044 $ & $ \bf 0.851 \pm 0.060 $ & $ \bf 0.853 \pm 0.060 $ & $ \bf 0.851 \pm 0.061 $ & $ \bf 0.846 \pm 0.058 $ & $ \bf 0.842 \pm 0.064 $ \\
            letter & \underline{$ \bf 0.665 \pm 0.005 $} & $ \bf 0.659 \pm 0.005 $ & $ 0.626 \pm 0.005 $ & $ 0.580 \pm 0.007 $ & $ 0.587 \pm 0.007 $ & $ 0.590 \pm 0.007 $ & $ 0.526 \pm 0.007 $ & $ 0.524 \pm 0.007 $ \\
            libras & $ \bf 0.564 \pm 0.027 $ & \underline{$ \bf 0.589 \pm 0.025 $} & $ \bf 0.549 \pm 0.024 $ & $ \bf 0.517 \pm 0.027 $ & $ \bf 0.519 \pm 0.026 $ & $ \bf 0.528 \pm 0.027 $ & $ 0.339 \pm 0.031 $ & $ 0.304 \pm 0.018 $ \\
            pendigits & \underline{$ \bf 0.797 \pm 0.006 $} & $ \bf 0.787 \pm 0.006 $ & $ 0.736 \pm 0.007 $ & $ 0.728 \pm 0.008 $ & $ 0.729 \pm 0.008 $ & $ 0.730 \pm 0.008 $ & $ 0.703 \pm 0.008 $ & $ 0.714 \pm 0.007 $ \\
            satimage & $ \bf 0.817 \pm 0.009 $ & \underline{$ \bf 0.825 \pm 0.008 $} & $ \bf 0.807 \pm 0.009 $ & $ 0.678 \pm 0.010 $ & $ 0.678 \pm 0.010 $ & $ 0.679 \pm 0.010 $ & $ 0.654 \pm 0.011 $ & $ 0.652 \pm 0.010 $ \\
            segment & $ \bf 0.876 \pm 0.011 $ & \underline{$ \bf 0.878 \pm 0.011 $} & $ \bf 0.855 \pm 0.012 $ & $ 0.780 \pm 0.015 $ & $ 0.780 \pm 0.014 $ & $ 0.781 \pm 0.014 $ & $ 0.758 \pm 0.014 $ & $ 0.760 \pm 0.014 $ \\
            vehicle & $ \bf 0.734 \pm 0.025 $ & \underline{$ \bf 0.750 \pm 0.025 $} & $ \bf 0.743 \pm 0.029 $ & $ 0.593 \pm 0.025 $ & $ 0.594 \pm 0.023 $ & $ 0.594 \pm 0.025 $ & $ 0.563 \pm 0.024 $ & $ 0.549 \pm 0.028 $ \\
            vowel & $ \bf 0.658 \pm 0.023 $ & \underline{$ \bf 0.669 \pm 0.024 $} & $ \bf 0.631 \pm 0.023 $ & $ 0.537 \pm 0.025 $ & $ 0.538 \pm 0.026 $ & $ 0.544 \pm 0.026 $ & $ 0.381 \pm 0.029 $ & $ 0.353 \pm 0.020 $ \\
            wine & $ \bf 0.811 \pm 0.055 $ & $ \bf 0.807 \pm 0.053 $ & \underline{$ \bf 0.814 \pm 0.054 $} & $ \bf 0.729 \pm 0.058 $ & $ \bf 0.730 \pm 0.057 $ & $ \bf 0.728 \pm 0.063 $ & $ \bf 0.724 \pm 0.057 $ & $ \bf 0.721 \pm 0.066 $ \\
            yeast & $ \bf 0.752 \pm 0.012 $ & \underline{$ \bf 0.759 \pm 0.010 $} & $ \bf 0.734 \pm 0.010 $ & $ 0.528 \pm 0.017 $ & $ 0.528 \pm 0.016 $ & $ 0.531 \pm 0.017 $ & $ 0.383 \pm 0.010 $ & $ 0.365 \pm 0.010 $ \\
            \bottomrule
            \end{tabular}
    }
\end{table}

\begin{table}[ht]
    \centering
    \caption{Average \(\tau_X\) score (mean and standard deviation) of algorithms with 60\% of missing positions for the synthetic datasets}
    \label{table:accuracy_60}
    \resizebox{\textwidth}{!}
    {
        \begin{tabular}{lrrrrrrrr}
            \toprule
            Dataset & Bucket pivot & Borda & Copeland & Markov chain 1 & Markov chain 2 & Markov chain 3 & Markov chain 4 & Maximal lottery \\
            \midrule
            authorship & $ \bf 0.676 \pm 0.035 $ & \underline{$ \bf 0.694 \pm 0.030 $} & $ \bf 0.693 \pm 0.027 $ & $ \bf 0.610 \pm 0.036 $ & $ \bf 0.610 \pm 0.035 $ & $ \bf 0.611 \pm 0.035 $ & $ \bf 0.606 \pm 0.034 $ & $ \bf 0.606 \pm 0.034 $ \\
            blocks & $ \bf 0.895 \pm 0.010 $ & \underline{$ \bf 0.915 \pm 0.008 $} & $ \bf 0.896 \pm 0.010 $ & $ \bf 0.893 \pm 0.009 $ & $ \bf 0.892 \pm 0.009 $ & $ \bf 0.894 \pm 0.008 $ & $ 0.873 \pm 0.011 $ & $ \bf 0.894 \pm 0.008 $ \\
            breast & $ \bf 0.548 \pm 0.092 $ & \underline{$ \bf 0.623 \pm 0.082 $} & $ \bf 0.593 \pm 0.095 $ & $ \bf 0.473 \pm 0.087 $ & $ \bf 0.471 \pm 0.091 $ & $ \bf 0.478 \pm 0.088 $ & $ \bf 0.482 \pm 0.078 $ & $ \bf 0.419 \pm 0.068 $ \\
            ecoli & $ \bf 0.566 \pm 0.048 $ & \underline{$ \bf 0.668 \pm 0.040 $} & $ \bf 0.578 \pm 0.033 $ & $ \bf 0.565 \pm 0.043 $ & $ \bf 0.562 \pm 0.041 $ & $ \bf 0.567 \pm 0.040 $ & $ 0.503 \pm 0.039 $ & $ \bf 0.528 \pm 0.040 $ \\
            glass & $ \bf 0.545 \pm 0.090 $ & \underline{$ \bf 0.632 \pm 0.056 $} & $ \bf 0.627 \pm 0.055 $ & $ \bf 0.449 \pm 0.053 $ & $ \bf 0.447 \pm 0.055 $ & $ \bf 0.453 \pm 0.052 $ & $ \bf 0.474 \pm 0.053 $ & $ 0.404 \pm 0.055 $ \\
            iris & $ \bf 0.820 \pm 0.097 $ & $ \bf 0.819 \pm 0.099 $ & \underline{$ \bf 0.838 \pm 0.078 $} & $ \bf 0.799 \pm 0.100 $ & $ \bf 0.799 \pm 0.099 $ & $ \bf 0.799 \pm 0.100 $ & $ \bf 0.799 \pm 0.092 $ & $ \bf 0.808 \pm 0.097 $ \\
            letter & $ \bf 0.633 \pm 0.005 $ & \underline{$ \bf 0.637 \pm 0.004 $} & $ 0.545 \pm 0.006 $ & $ 0.559 \pm 0.009 $ & $ 0.570 \pm 0.009 $ & $ 0.575 \pm 0.009 $ & $ 0.514 \pm 0.010 $ & $ 0.526 \pm 0.007 $ \\
            libras & $ \bf 0.470 \pm 0.039 $ & \underline{$ \bf 0.535 \pm 0.033 $} & $ \bf 0.490 \pm 0.031 $ & $ \bf 0.422 \pm 0.028 $ & $ \bf 0.431 \pm 0.029 $ & $ \bf 0.440 \pm 0.030 $ & $ 0.392 \pm 0.034 $ & $ 0.302 \pm 0.025 $ \\
            pendigits & $ 0.730 \pm 0.008 $ & \underline{$ \bf 0.758 \pm 0.006 $} & $ 0.633 \pm 0.008 $ & $ 0.711 \pm 0.009 $ & $ 0.712 \pm 0.008 $ & $ 0.714 \pm 0.008 $ & $ 0.598 \pm 0.010 $ & $ 0.705 \pm 0.008 $ \\
            satimage & $ 0.728 \pm 0.013 $ & \underline{$ \bf 0.779 \pm 0.009 $} & $ \bf 0.748 \pm 0.010 $ & $ 0.648 \pm 0.012 $ & $ 0.647 \pm 0.011 $ & $ 0.651 \pm 0.012 $ & $ 0.628 \pm 0.012 $ & $ 0.643 \pm 0.010 $ \\
            segment & $ \bf 0.797 \pm 0.017 $ & \underline{$ \bf 0.836 \pm 0.011 $} & $ 0.779 \pm 0.016 $ & $ 0.756 \pm 0.014 $ & $ 0.755 \pm 0.016 $ & $ 0.758 \pm 0.015 $ & $ 0.711 \pm 0.021 $ & $ 0.752 \pm 0.015 $ \\
            vehicle & $ \bf 0.627 \pm 0.034 $ & \underline{$ \bf 0.663 \pm 0.033 $} & $ \bf 0.661 \pm 0.033 $ & $ \bf 0.524 \pm 0.041 $ & $ \bf 0.524 \pm 0.043 $ & $ \bf 0.525 \pm 0.042 $ & $ \bf 0.514 \pm 0.045 $ & $ 0.497 \pm 0.042 $ \\
            vowel & $ \bf 0.521 \pm 0.031 $ & \underline{$ \bf 0.603 \pm 0.024 $} & $ \bf 0.556 \pm 0.026 $ & $ 0.426 \pm 0.039 $ & $ 0.429 \pm 0.038 $ & $ 0.434 \pm 0.038 $ & $ 0.436 \pm 0.023 $ & $ 0.334 \pm 0.024 $ \\
            wine & $ \bf 0.684 \pm 0.103 $ & $ \bf 0.680 \pm 0.103 $ & \underline{$ \bf 0.701 \pm 0.100 $} & $ \bf 0.627 \pm 0.088 $ & $ \bf 0.628 \pm 0.088 $ & $ \bf 0.628 \pm 0.088 $ & $ \bf 0.624 \pm 0.098 $ & $ \bf 0.612 \pm 0.099 $ \\
            yeast & $ 0.578 \pm 0.021 $ & \underline{$ \bf 0.710 \pm 0.012 $} & $ \bf 0.676 \pm 0.012 $ & $ 0.428 \pm 0.019 $ & $ 0.430 \pm 0.019 $ & $ 0.432 \pm 0.019 $ & $ 0.484 \pm 0.017 $ & $ 0.362 \pm 0.011 $ \\
            \bottomrule
            \end{tabular}
        }
\end{table}

\clearpage
\newpage

\section{Complete training time results}

This section presents the complete training time results for all aggregation algorithms evaluated across 16 synthetic datasets under varying percentages of missing class labels: 0\%, 30\%, and 60\%. The reported values correspond to the average training time in seconds, measured over multiple runs, with standard deviations provided in parentheses. In each table, the \underline{underlined} values indicate the fastest algorithm(s) for a given dataset. \textbf{Boldfaced} values highlight algorithms whose 95\% confidence intervals, defined by the mean and standard deviation, overlap with that of the fastest algorithm.

\begin{table}[ht]
    \centering
    \caption{Average training time (in seconds, with standard deviation) of algorithms with 0\% of missing class label positions for the synthetic datasets}
    \resizebox{\textwidth}{!}
    {
        \begin{tabular}{lrrrrrrrr}
            \toprule
            Dataset & Bucket pivot & Borda & Copeland & Markov chain 1 & Markov chain 2 & Markov chain 3 & Markov chain 4 & Maximal lottery \\
            \midrule
            authorship & $ 0.678 \pm 0.024 $ & \underline{$\bf 0.341 \pm 0.013 $} & \underline{$\bf 0.341 \pm 0.013 $} & $ 0.965 \pm 0.039 $ & $ 0.955 \pm 0.027 $ & $ 0.923 \pm 0.022 $ & $ 0.557 \pm 0.020 $ & $ 6.979 \pm 0.344 $ \\
            blocks & $ 1.232 \pm 0.050 $ & $\bf 0.697 \pm 0.034 $ & \underline{$\bf 0.695 \pm 0.031 $} & $ 2.236 \pm 0.075 $ & $ 2.247 \pm 0.070 $ & $ 2.239 \pm 0.063 $ & $ 1.409 \pm 0.053 $ & $ 14.068 \pm 0.659 $ \\
            breast & $ 0.027 \pm 0.003 $ & \underline{$\bf 0.014 \pm 0.002 $} & \underline{$\bf 0.014 \pm 0.002 $} & $ 0.067 \pm 0.006 $ & $ 0.066 \pm 0.004 $ & $ 0.066 \pm 0.005 $ & $ 0.031 \pm 0.002 $ & $ 0.752 \pm 0.148 $ \\
            ecoli & $ 0.045 \pm 0.004 $ & \underline{$\bf 0.027 \pm 0.002 $} & $\bf 0.028 \pm 0.003 $ & $ 0.111 \pm 0.009 $ & $ 0.114 \pm 0.010 $ & $ 0.110 \pm 0.005 $ & $ 0.055 \pm 0.004 $ & $ 0.679 \pm 0.071 $ \\
            glass & $ 0.051 \pm 0.004 $ & \underline{$\bf 0.024 \pm 0.003 $} & \underline{$\bf 0.024 \pm 0.002 $} & $ 0.108 \pm 0.007 $ & $ 0.109 \pm 0.007 $ & $ 0.110 \pm 0.006 $ & $ 0.051 \pm 0.004 $ & $ 0.848 \pm 0.089 $ \\
            iris & $\bf 0.007 \pm 0.002 $ & \underline{$\bf 0.005 \pm 0.002 $} & \underline{$\bf 0.005 \pm 0.002 $} & $\bf 0.011 \pm 0.003 $ & $\bf 0.009 \pm 0.002 $ & $\bf 0.009 \pm 0.002 $ & $\bf 0.007 \pm 0.002 $ & $ 0.266 \pm 0.057 $ \\
            letter & $ 18.073 \pm 0.119 $ & \underline{$\bf 17.501 \pm 0.134 $} & $\bf 17.554 \pm 0.137 $ & $ 30.068 \pm 0.231 $ & $ 29.979 \pm 0.240 $ & $ 30.114 \pm 0.225 $ & $ 24.007 \pm 0.194 $ & $ 46.851 \pm 1.981 $ \\
            libras & $ 2.121 \pm 0.063 $ & \underline{$\bf 1.640 \pm 0.047 $} & $\bf 1.685 \pm 0.044 $ & $ 9.650 \pm 0.214 $ & $ 9.643 \pm 0.221 $ & $ 9.718 \pm 0.230 $ & $ 4.654 \pm 0.113 $ & $ 29.088 \pm 1.044 $ \\
            pendigits & $ 4.848 \pm 0.060 $ & \underline{$\bf 3.433 \pm 0.044 $} & $\bf 3.434 \pm 0.049 $ & $ 13.308 \pm 0.153 $ & $ 13.567 \pm 0.140 $ & $ 13.598 \pm 0.149 $ & $ 7.249 \pm 0.081 $ & $ 41.310 \pm 1.269 $ \\
            satimage & $ 3.527 \pm 0.063 $ & \underline{$\bf 2.050 \pm 0.039 $} & $\bf 2.053 \pm 0.039 $ & $ 6.412 \pm 0.090 $ & $ 6.475 \pm 0.101 $ & $ 6.456 \pm 0.086 $ & $ 3.622 \pm 0.054 $ & $ 32.148 \pm 0.633 $ \\
            segment & $ 1.300 \pm 0.033 $ & \underline{$\bf 0.682 \pm 0.017 $} & $\bf 0.689 \pm 0.021 $ & $ 3.530 \pm 0.099 $ & $ 3.576 \pm 0.074 $ & $ 3.618 \pm 0.070 $ & $ 1.744 \pm 0.045 $ & $ 18.949 \pm 1.128 $ \\
            vehicle & $ 0.228 \pm 0.013 $ & \underline{$\bf 0.118 \pm 0.009 $} & $\bf 0.122 \pm 0.009 $ & $ 0.352 \pm 0.015 $ & $ 0.371 \pm 0.025 $ & $ 0.349 \pm 0.015 $ & $ 0.211 \pm 0.020 $ & $ 2.999 \pm 0.125 $ \\
            vowel & $ 0.338 \pm 0.012 $ & \underline{$\bf 0.230 \pm 0.009 $} & $\bf 0.234 \pm 0.011 $ & $ 1.365 \pm 0.047 $ & $ 1.363 \pm 0.038 $ & $ 1.387 \pm 0.046 $ & $ 0.559 \pm 0.019 $ & $ 5.749 \pm 0.385 $ \\
            wine & $ 0.041 \pm 0.004 $ & \underline{$\bf 0.018 \pm 0.002 $} & $\bf 0.020 \pm 0.006 $ & $ 0.056 \pm 0.004 $ & $ 0.055 \pm 0.004 $ & $ 0.055 \pm 0.003 $ & $ 0.033 \pm 0.002 $ & $ 0.905 \pm 0.118 $ \\
            yeast & $ 0.256 \pm 0.011 $ & \underline{$\bf 0.161 \pm 0.010 $} & $\bf 0.162 \pm 0.008 $ & $ 0.603 \pm 0.029 $ & $ 0.603 \pm 0.026 $ & $ 0.617 \pm 0.023 $ & $ 0.295 \pm 0.013 $ & $ 2.093 \pm 0.162 $ \\
            \bottomrule
        \end{tabular}
    }
\end{table}

\begin{table}[ht]
    \centering
    \caption{Average training time (in seconds, with standard deviation) of algorithms with 30\% of missing class label positions for the synthetic datasets}
    \resizebox{\textwidth}{!}
    {
        \begin{tabular}{lrrrrrrrr}
            \toprule
            Dataset & Bucket pivot & Borda & Copeland & Markov chain 1 & Markov chain 2 & Markov chain 3 & Markov chain 4 & Maximal lottery \\
            \midrule
            authorship & $ 0.537 \pm 0.021 $ & $\bf 0.294 \pm 0.019 $ & \underline{$\bf 0.282 \pm 0.009 $} & $ 0.784 \pm 0.029 $ & $ 0.797 \pm 0.031 $ & $ 0.784 \pm 0.040 $ & $ 0.516 \pm 0.017 $ & $ 5.731 \pm 0.217 $ \\
            blocks & $ 1.142 \pm 0.065 $ & \underline{$\bf 0.638 \pm 0.038 $} & $\bf 0.646 \pm 0.043 $ & $ 2.230 \pm 0.119 $ & $ 2.229 \pm 0.112 $ & $ 2.227 \pm 0.117 $ & $ 1.443 \pm 0.082 $ & $ 13.328 \pm 0.813 $ \\
            breast & $ 0.023 \pm 0.002 $ & \underline{$\bf 0.012 \pm 0.002 $} & $\bf 0.013 \pm 0.001 $ & $ 0.064 \pm 0.006 $ & $ 0.064 \pm 0.007 $ & $ 0.061 \pm 0.004 $ & $ 0.035 \pm 0.003 $ & $ 0.700 \pm 0.076 $ \\
            ecoli & $ 0.041 \pm 0.003 $ & \underline{$\bf 0.025 \pm 0.002 $} & \underline{$\bf 0.025 \pm 0.002 $} & $ 0.113 \pm 0.008 $ & $ 0.115 \pm 0.008 $ & $ 0.110 \pm 0.007 $ & $ 0.068 \pm 0.007 $ & $ 0.646 \pm 0.058 $ \\
            glass & $ 0.047 \pm 0.004 $ & \underline{$\bf 0.022 \pm 0.003 $} & $\bf 0.023 \pm 0.002 $ & $ 0.108 \pm 0.008 $ & $ 0.107 \pm 0.007 $ & $ 0.105 \pm 0.008 $ & $ 0.057 \pm 0.004 $ & $ 0.855 \pm 0.143 $ \\
            iris & $\bf 0.005 \pm 0.001 $ & \underline{$\bf 0.004 \pm 0.001 $} & \underline{$\bf 0.004 \pm 0.001 $} & $\bf 0.007 \pm 0.002 $ & $\bf 0.007 \pm 0.002 $ & $\bf 0.007 \pm 0.002 $ & $\bf 0.006 \pm 0.001 $ & $ 0.252 \pm 0.030 $ \\
            letter & $\bf 14.820 \pm 0.104 $ & \underline{$\bf 14.454 \pm 0.100 $} & $\bf 14.517 \pm 0.112 $ & $ 28.011 \pm 0.249 $ & $ 27.955 \pm 0.249 $ & $ 27.872 \pm 0.251 $ & $ 24.116 \pm 0.233 $ & $ 39.714 \pm 1.003 $ \\
            libras & $ 2.004 \pm 0.080 $ & \underline{$\bf 1.536 \pm 0.059 $} & $\bf 1.589 \pm 0.059 $ & $ 9.828 \pm 0.368 $ & $ 9.777 \pm 0.358 $ & $ 9.708 \pm 0.378 $ & $ 6.215 \pm 0.268 $ & $ 29.244 \pm 1.233 $ \\
            pendigits & $ 4.810 \pm 0.061 $ & \underline{$\bf 3.498 \pm 0.049 $} & $\bf 3.512 \pm 0.052 $ & $ 14.341 \pm 0.157 $ & $ 14.450 \pm 0.183 $ & $ 14.224 \pm 0.132 $ & $ 9.224 \pm 0.100 $ & $ 41.717 \pm 0.892 $ \\
            satimage & $ 3.151 \pm 0.058 $ & $\bf 1.895 \pm 0.039 $ & \underline{$\bf 1.887 \pm 0.030 $} & $ 6.169 \pm 0.104 $ & $ 6.232 \pm 0.084 $ & $ 6.137 \pm 0.075 $ & $ 3.899 \pm 0.085 $ & $ 29.956 \pm 1.192 $ \\
            segment & $ 1.199 \pm 0.038 $ & \underline{$\bf 0.635 \pm 0.017 $} & $\bf 0.640 \pm 0.020 $ & $ 3.563 \pm 0.108 $ & $ 3.611 \pm 0.099 $ & $ 3.568 \pm 0.110 $ & $ 1.892 \pm 0.071 $ & $ 18.501 \pm 0.657 $ \\
            vehicle & $ 0.192 \pm 0.010 $ & \underline{$\bf 0.101 \pm 0.007 $} & $\bf 0.104 \pm 0.008 $ & $ 0.312 \pm 0.015 $ & $ 0.318 \pm 0.018 $ & $ 0.312 \pm 0.015 $ & $ 0.195 \pm 0.010 $ & $ 2.679 \pm 0.123 $ \\
            vowel & $ 0.331 \pm 0.013 $ & \underline{$\bf 0.215 \pm 0.009 $} & $\bf 0.224 \pm 0.008 $ & $ 1.407 \pm 0.044 $ & $ 1.395 \pm 0.037 $ & $ 1.389 \pm 0.039 $ & $ 0.728 \pm 0.030 $ & $ 5.642 \pm 0.262 $ \\
            wine & $ 0.030 \pm 0.003 $ & $\bf 0.015 \pm 0.003 $ & \underline{$\bf 0.014 \pm 0.002 $} & $ 0.042 \pm 0.004 $ & $ 0.042 \pm 0.004 $ & $ 0.042 \pm 0.004 $ & $ 0.028 \pm 0.003 $ & $ 0.752 \pm 0.085 $ \\
            yeast & $ 0.256 \pm 0.016 $ & \underline{$\bf 0.166 \pm 0.012 $} & $\bf 0.167 \pm 0.009 $ & $ 0.624 \pm 0.031 $ & $ 0.623 \pm 0.022 $ & $ 0.629 \pm 0.024 $ & $ 0.362 \pm 0.017 $ & $ 2.004 \pm 0.090 $ \\
            \bottomrule
        \end{tabular}
    }
\end{table}

\begin{table}[ht]
    \centering
    \caption{Average training time (in seconds, with standard deviation) of algorithms with 60\% of missing class label positions for the synthetic datasets}
    \resizebox{\textwidth}{!}
    {
        \begin{tabular}{lrrrrrrrr}
            \toprule
            Dataset & Bucket pivot & Borda & Copeland & Markov chain 1 & Markov chain 2 & Markov chain 3 & Markov chain 4 & Maximal lottery \\
            \midrule
            authorship & $ 0.317 \pm 0.022 $ & \underline{$\bf 0.186 \pm 0.013 $} & \underline{$\bf 0.186 \pm 0.010 $} & $ 0.481 \pm 0.028 $ & $ 0.491 \pm 0.033 $ & $ 0.477 \pm 0.028 $ & $ 0.350 \pm 0.021 $ & $ 3.633 \pm 0.325 $ \\
            blocks & $ 0.852 \pm 0.061 $ & \underline{$\bf 0.480 \pm 0.034 $} & $\bf 0.481 \pm 0.036 $ & $ 1.733 \pm 0.118 $ & $ 1.739 \pm 0.115 $ & $ 1.723 \pm 0.119 $ & $ 1.191 \pm 0.087 $ & $ 10.469 \pm 0.665 $ \\
            breast & $ 0.018 \pm 0.002 $ & \underline{$\bf 0.009 \pm 0.001 $} & $\bf 0.010 \pm 0.001 $ & $ 0.054 \pm 0.005 $ & $ 0.054 \pm 0.005 $ & $ 0.052 \pm 0.005 $ & $ 0.041 \pm 0.004 $ & $ 0.611 \pm 0.102 $ \\
            ecoli & $ 0.032 \pm 0.003 $ & \underline{$\bf 0.020 \pm 0.003 $} & \underline{$\bf 0.020 \pm 0.002 $} & $ 0.098 \pm 0.007 $ & $ 0.101 \pm 0.007 $ & $ 0.097 \pm 0.006 $ & $ 0.076 \pm 0.005 $ & $ 0.606 \pm 0.049 $ \\
            glass & $ 0.034 \pm 0.004 $ & \underline{$\bf 0.017 \pm 0.002 $} & \underline{$\bf 0.017 \pm 0.002 $} & $ 0.085 \pm 0.007 $ & $ 0.089 \pm 0.008 $ & $ 0.086 \pm 0.006 $ & $ 0.060 \pm 0.005 $ & $ 0.751 \pm 0.106 $ \\
            iris & $\bf 0.004 \pm 0.001 $ & $\bf 0.004 \pm 0.002 $ & \underline{$\bf 0.003 \pm 0.001 $} & $\bf 0.005 \pm 0.001 $ & $\bf 0.005 \pm 0.001 $ & $\bf 0.005 \pm 0.001 $ & $\bf 0.004 \pm 0.001 $ & $ 0.230 \pm 0.032 $ \\
            letter & $\bf 9.771 \pm 0.107 $ & \underline{$\bf 9.409 \pm 0.112 $} & $\bf 9.539 \pm 0.088 $ & $ 22.307 \pm 0.190 $ & $ 22.322 \pm 0.204 $ & $ 22.089 \pm 0.156 $ & $ 20.703 \pm 0.186 $ & $ 34.987 \pm 0.854 $ \\
            libras & $ 1.521 \pm 0.047 $ & \underline{$\bf 1.135 \pm 0.044 $} & $\bf 1.183 \pm 0.036 $ & $ 9.062 \pm 0.253 $ & $ 9.072 \pm 0.251 $ & $ 8.855 \pm 0.244 $ & $ 7.830 \pm 0.252 $ & $ 28.221 \pm 1.337 $ \\
            pendigits & $ 3.553 \pm 0.036 $ & \underline{$\bf 2.578 \pm 0.043 $} & $\bf 2.662 \pm 0.082 $ & $ 12.224 \pm 0.122 $ & $ 12.420 \pm 0.105 $ & $ 12.178 \pm 0.095 $ & $ 9.359 \pm 0.091 $ & $ 35.583 \pm 1.162 $ \\
            satimage & $ 2.235 \pm 0.035 $ & \underline{$\bf 1.409 \pm 0.028 $} & $\bf 1.424 \pm 0.033 $ & $ 4.729 \pm 0.095 $ & $ 4.821 \pm 0.071 $ & $ 4.684 \pm 0.080 $ & $ 3.475 \pm 0.066 $ & $ 23.182 \pm 0.977 $ \\
            segment & $ 0.932 \pm 0.036 $ & \underline{$\bf 0.493 \pm 0.015 $} & $\bf 0.499 \pm 0.017 $ & $ 3.064 \pm 0.113 $ & $ 3.081 \pm 0.100 $ & $ 3.040 \pm 0.099 $ & $ 1.878 \pm 0.072 $ & $ 16.602 \pm 1.004 $ \\
            vehicle & $ 0.122 \pm 0.006 $ & \underline{$\bf 0.068 \pm 0.004 $} & $\bf 0.069 \pm 0.004 $ & $ 0.210 \pm 0.011 $ & $ 0.212 \pm 0.012 $ & $ 0.209 \pm 0.013 $ & $ 0.145 \pm 0.009 $ & $ 1.973 \pm 0.135 $ \\
            vowel & $ 0.268 \pm 0.009 $ & \underline{$\bf 0.171 \pm 0.007 $} & $\bf 0.178 \pm 0.007 $ & $ 1.323 \pm 0.037 $ & $ 1.327 \pm 0.035 $ & $ 1.303 \pm 0.038 $ & $ 0.966 \pm 0.038 $ & $ 5.533 \pm 0.313 $ \\
            wine & $ 0.018 \pm 0.002 $ & \underline{$\bf 0.009 \pm 0.002 $} & \underline{$\bf 0.009 \pm 0.002 $} & $ 0.027 \pm 0.004 $ & $ 0.027 \pm 0.003 $ & $ 0.027 \pm 0.003 $ & $ 0.019 \pm 0.002 $ & $ 0.572 \pm 0.066 $ \\
            yeast & $ 0.191 \pm 0.009 $ & \underline{$\bf 0.123 \pm 0.006 $} & $\bf 0.125 \pm 0.004 $ & $ 0.542 \pm 0.023 $ & $ 0.556 \pm 0.013 $ & $ 0.535 \pm 0.013 $ & $ 0.396 \pm 0.015 $ & $ 1.877 \pm 0.139 $ \\
            \bottomrule
        \end{tabular}
    }
\end{table}

\clearpage
\newpage

\section{Complete bucket count difference results}

This section presents the complete results for the discrepancy between the number of buckets in the predicted and true bucket orders across all datasets, aggregation algorithms, and levels of missing class labels.

\begin{figure}[ht]
    \centering
    \includegraphics[width = 13.25cm]{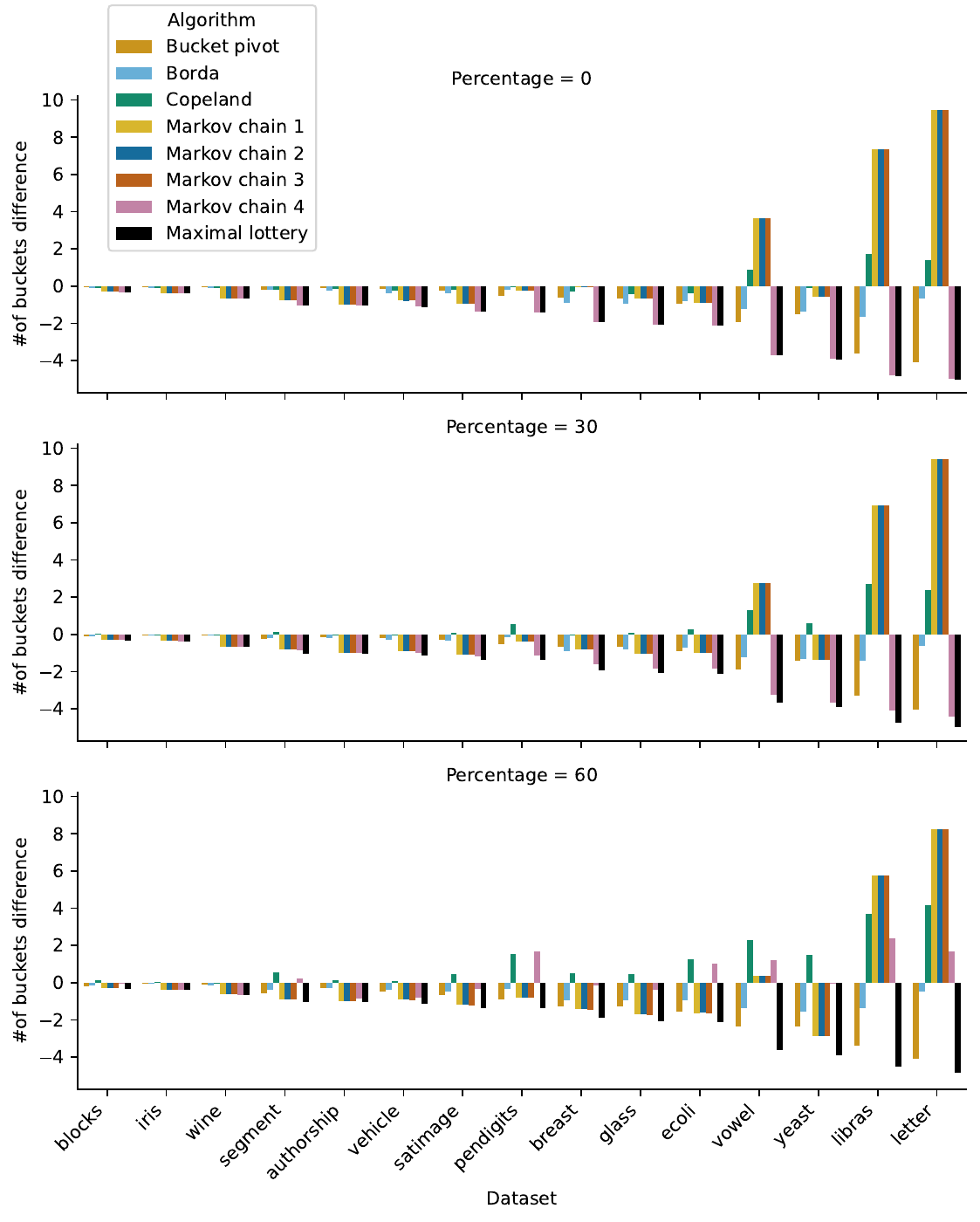}
    \caption{Average bucket count difference between the true and predicted bucket orders for all aggregation algorithms across all datasets and missing label settings.}
    \label{fig:bucket_diff_all}
\end{figure}
\end{document}